\documentclass[number,preprint,12pt]{elsarticle}
\usepackage{amssymb}
\usepackage[numbers]{natbib}
\usepackage{hyperref}
\usepackage{subfigure}
\usepackage{threeparttable}
\usepackage{booktabs}
\usepackage{amsmath}
\usepackage{multirow}
\usepackage{graphicx}
\usepackage{tabularx}
\usepackage{subfigure}
\usepackage{adjustbox}

\usepackage{fancyhdr}

\makeatletter
\def\bibinfo#1{%
  \@ifundefined{bibinfo@X@#1}%
    {\@firstofone}
    {\csname bibinfo@X@#1\endcsname}}
\makeatother

\fancypagestyle{pprintTitle}{%
\lhead{\small \textbf{This article has been submitted to Elsevier for possible publication. Copyright may be transferred without notice and this version might not be longer available}\\} \chead{}\rhead{\scriptsize}
\lfoot{\small \textit{Preprint submitted to Computer Networks}}\cfoot{}\rfoot{\small \textit{\today}}

\setlength{\headheight}{52.4842pt}
\addtolength{\topmargin}{-20.4842pt}
}
\journal{Computer Networks}
\begin{document}

\begin{frontmatter}

\title{Measuring and Estimating Key Quality Indicators in Cloud Gaming services.}

\affiliation[inst1]{organization={Telecommunication Research Institute (TELMA), Universidad de Málaga},%Department and Organization
            addressline={E.T.S. Ingeniería de Telecomunicación, Bulevar Louis Pasteur 35}, 
            city={Malaga},
            postcode={29010}, 
            state={Andalucia},
            country={Spain}}

\author[inst1]{Carlos Baena}
\ead{jcbg@ic.uma.es}
\author[inst1]{O. S. Peñaherrera-Pulla}
\ead{sppulla@ic.uma.es}
\author[inst1]{Raquel Barco}
\ead{rbm@ic.uma.es}
\author[inst1]{Sergio Fortes\corref{cor1}}
\ead{sfr@ic.uma.es}

\cortext[cor1]{ Corresponding author}
\fntext[label2]{ This work has been partially funded by: Ministerio de Asuntos Económicos y Transformación Digital and European Union - NextGenerationEU within the framework ``Recuperación, Transformación y Resiliencia y el Mecanismo de Recuperación y Resiliencia'' under the project MAORI, and Universidad de Málaga through the ``II Plan Propio de Investigación, Transferencia y Divulgación Científica''. 
This work has been also supported by Junta de Andalucía through Secretaría General de Universidades, Investigación y Tecnología with the predoctoral grant (Ref. PREDOC\_01712) as well as by Ministerio de Ciencia y Tecnología through grant FPU19/04468.}

\begin{abstract}

User equipment is one of the main bottlenecks facing the gaming industry nowadays. 
The extremely realistic games which are currently available trigger high computational requirements of the user devices to run games.
As a consequence, the game industry has proposed the concept of Cloud Gaming, a paradigm that improves gaming experience in reduced hardware devices.
To this end, games are hosted on remote servers, relegating users’ devices to play only the role of a peripheral for interacting with the game.
However, this paradigm overloads the communication links connecting the users with the cloud.
Therefore, service experience becomes highly dependent on network connectivity. To overcome this, Cloud Gaming will be boosted by the promised performance of 5G and future 6G networks, together with the flexibility provided by mobility in multi-RAT scenarios, such as WiFi.
In this scope, the present work proposes a framework for measuring and estimating the main E2E metrics of the Cloud Gaming service, namely KQIs.
In addition, different machine learning techniques are assessed for predicting KQIs related to Cloud Gaming user's experience.
To this end, the main key quality indicators (KQIs) of the service such as input lag, freeze percent or perceived video frame rate are collected in a real environment.
Based on these, results show that machine learning techniques provide a good estimation of these indicators solely from network-based metrics. This is considered a valuable asset to guide the delivery of Cloud Gaming services through cellular communications networks  even without access to the user's device, as it is expected for telecom operators.

\end{abstract}

\iffalse
\begin{graphicalabstract}
\includegraphics{grabs}
\end{graphicalabstract}
\fi

\iffalse
\begin{highlights}
\item Novel framework for Cloud Gaming E2E metrics acquisition and estimation. 
\item Different ML approaches are assessed through the estimation error and prediction time, also considering the source and number of estimators.
\item ML models generally shows remarkable estimation of the KQI, generally requiring about four predictors to achieve a good performance.
\item RF and KNR are positioned as the most interesting techniques for the estimation of Cloud Gaming KQIs such as E2E latency, freeze percent or the frame rate perceived by the user.
\item This ML-based framework will enable intelligent network management for new-generation mobile networks thanks to its potential integration on the Open RAN paradigm through the RIC.
\end{highlights}

\fi

\end{frontmatter}

\section{Introduction}
\label{section:introduction}
The entertainment sector has recently experimented a huge growth thanks to multimedia services like gaming, whose industry has been positioned  in the last few years as one of the top profit sources in the recreational area \cite{MARCHAND2013141}.
This industry has been able to catch people's attention by extending the original goal pursued since its foundation: offering a virtual world where players can perform a myriad of actions as well as interacting \cite{Ducheneaut} and socialising with other users \cite{SZELL2010313}.

These virtual worlds are increasingly closer to reality, offering from sharper 2D environments to 3D augmented and virtual scenarios. Nonetheless, presenting to the player these environments requires computationally expensive tasks, forcing users to invest a large amount of resources in devices able to introduce them to this kind of realistic environments.

Therefore, game industry is trying to take advantage of the high capabilities of cloud computing via the Cloud Gaming (CG) paradigm \cite{cai2016survey}. This concept aims to move the execution of rendering tasks to the cloud, converting user devices into thin clients. In this way, the only role of the user equipment is to be the interface between users and the virtual world, gathering users' input actions and displaying the video scenes generated by the remote server where the environment is rendered.

In this way, users are allowed to play nearly any game at any place at any moment on any device, since decoding tasks are supported by the vast mass-produced chips which are included in the cheapest market devices. 
Moreover, this paradigm enables the introduction of Game on Demand (GoD), which avoids the installation of any game in users' equipment and enables new ways of commercialization (e.g., via subscription). 

When it comes to game companies, CG implies a strong asset for combating piracy, since a game copy will never be downloaded in users' devices. Additionally, this concept eases platform compatibility issues, which might also reduce game production costs.

As its main drawback, with CG the service becomes totally dependent on the network. Conversely, traditional games only required a simple network connection to enable online multiplayer, allowing users to interact, despite their geographical location.
In this sense, the information which is usually sent in classic online games ranges from player's action and position to the attire or object that each character carries at each moment. Nonetheless, this information typically requires only the exchange of small packets. 

Conversely, in CG the fact of rendering the environment in a remote server leads to a huge increment in network traffic demand: the demanding latency requirements hinder the application of content compression techniques, increasing the volume of data in comparison with traditional video services.
In addition, the interactive essence of video games makes this paradigm be more sensible to network latency than other streaming services \cite{Abdallah2018}. Consequently, the network is the most critical element in CG performance.

In this context, the Fifth mobile generation (5G) networks can boost CG services.
The high data rates and low latency values expected with the service categories introduced and supported by 5G (i.e., enhanced Mobile Broadband (eMBB) and Ultra Reliability and Low Latency Communications (URLLC)) places this new technology as the key to enabling CG paradigm \cite{BaenaCG2022}.

This fact, together with the improvements in cloud infrastructure, has attracted the interest of many of the largest tech companies, which have started their own CG services, such as Playstation Now (Sony) \cite{playstation}, Nvidia GeForceNow \cite{geforcenow}, Amazon Luna \cite{amazonLuna} or Microsoft xCloud \cite{xbox}. Simultaneously, telecom operators have identified these platforms as a potential way to offer exclusive services to their clients, being a possible key differentiating factor in the market.

However, network management becomes extremely complex in the scenarios introduced by 5G. Consequently, the use of self-organising networks (SON) algorithms together with machine learning (ML) techniques are highlighted as the prime solution to cover these tasks \cite{MLSON, UMLNetworking, Palacios2018, LocationAwareness}. 
Together with SON, network operators start to follow a new management strategy based on quality of experience (QoE). In this trend, network is configured and optimised based on service key quality indicators (KQIs) \cite{herrera2019modeling, baena2020estimation}, pursuing to improve the user's perceived quality of service. Nevertheless, the collection of these E2E metrics supposes a challenging task. On the one hand, accessing device data is generally limited, as well as supposes a potential threat to the confidentiality of user data. On the other hand, the increasingly adoption of security protocols hinders their calculation from traditional techniques such as packet inspection.

In this context, the contributions of this paper are as follows. 
First, CG QoE measurement is analysed, highlighting the most paramount indicators to quantify it. 
Based on this, a novel framework to measure these KQIs is proposed. This framework allows obtaining this kind of metric without the need to access the application data.
Moreover, an ML-based approach is presented to predict these KQIs from common cellular network metrics. Finally, an evaluation of key ML regression techniques is performed assessing their accuracy and prediction time for the different CG KQIs.

The rest of the present article is organised as follows. Section \ref{sec:RelatedWorks} presents related works and exposes the main contributions of this work. Section \ref{sec:cgQualityInd}
describes the main KQIs for CG services. Then, Section \ref{sec:dataAcquisiton} provides the methodology followed in the framework for gathering these metrics. Thanks to its use, an analysis of these metrics is carried out in a real CG scenario. Section \ref{sec:estimationMethod} exposes the estimation part of the framework. Subsequently, Section \ref{sec:evaluation} provides an assessment of the key ML regression techniques considered. Finally, Section \ref{sec:Conclusions} shows the conclusions and future research lines.

\section{Related work and contributions}
\label{sec:RelatedWorks}

Since interaction is the main attraction of games, the Quality of Experience (QoE) of Online and CG is highly dependent on system response, thus on network latency.
This has been extensively covered in the literature.
In \cite{jarschel2013gaming}, authors present a MOS study in which  they expose the high impact of network latency and packet loss on QoE of CG. Following this line, authors in \cite{claypool2014effects} show the lineal correlation between network latency and MOS degradation by means of both an objective and subjective study. 
Raeen et al \cite{Raaen2014} expose through their study that the majority of gamers are not able to differentiate response times below 40 ms. However they also indicate that about half the occasional users can not tolerate service response values over 100ms.

In the same line, authors in \cite{Sabet2019DelayVariation} study how gamers adapt to different variations on delay using both objective and subjective methodologies. Results show that users can adapt to a constant delay, whereas frequent delay switching annoys gamers.
Besides, in \cite{Sabet2020} the influence of user strategy on the delay sensitivity is studied, concluding that the impact of delay sensitivity  on QoE has no correlation with the strategy chosen by the gamer to complete a task.

Other works such as \cite{claypool2006latency}, \cite{quax2013evaluation} and \cite{claypool2010lat} have studied  the impact that network latency presents on these services depending on the game type, e.g., action, puzzle, etc.
Hence, focusing on online games, authors in \cite{claypool2006latency} and \cite{claypool2010lat} show the importance of latency in the user performance, suggesting that latency perception is determined by the precision and deadline of the game action. Actions with high  precision and tight deadlines (e.g., first-person shooter) are more sensitive to latency than those that require low precision and do not have immediate deadlines (e.g., strategy games).

Using OnLive CG platform, Quax et al. \cite{quax2013evaluation} provide a qualitative comparison among action, strategy, puzzle, and racing types, pointing to action-oriented games as the most critical in terms of latency.
Likewise, authors in \cite{lee2012all} develop a model to predict different game strictness based on the rate of players' inputs and the game screen dynamics, easing the detection of a CG's latency sensibility. An extended prediction model is provided in \cite{Sabet2020DelaySensityClassification}, where the authors use other game characteristics such as Temporal and Spatial Accuracy, Degree of Freedom, Consequences, Importance of Actions, number or required actions among others.

Conversely,  Slivar et al \cite{slivar2015} shows through an empirical QoE study that CG services are more sensitive to network conditions than Online Gaming. By testing both paradigms under different network conditions, they conclude that CG services suffer higher degradation than a traditional game architecture in terms of user experience. Moreover, they highlight the importance of ensuring adequate video quality.
This latter fact is taken into account in the MOS study carried out by authors in \cite{moller2013factors}, in which video quality is positioned together with input sensitivity as a primordial feature to calculate CG QoE.

Authors in \cite{chang2011understanding} assess the performance of some thin clients such as UltraVNC and TeamViewer within the CG concept. Based on the graphic quality as well as the image fluency, they show that this kind of widely extended platform for remote desktop purposes is not able to support CG, since low values of frame rate are reached.
Additionally, Claypool et al. \cite{Claypool} present the important role that the network plays in streaming parameters through their study of CG traffic features. Based on bitrate, frequency and volume of data, their study shows that a poor network quality (in terms of high packet loss rate and insufficient bandwidth) leads to poor quality of the service.

In this context, network operators pursue the provision of CG services with the best QoE possible. 
Nonetheless, the complex architecture presented by the latest cellular networks, together with the provision on the same infrastructure of multiple services with heterogeneous requirements, make network management tasks extremely complex. In this sense,  optimised network management is becoming increasingly important to offer a good service.

As a solution, machine learning (ML) techniques have been recently proposed to support management tasks.
Indeed, Herrera et al. \cite{herrera2019modeling} focus on FTP (File-Transport-Protocol) services to present an useful KQI modeling for the management of new-generation mobile networks.
Along the same line, authors in \cite{baena2020estimation} suggest a system based on video KQI estimation to support network slicing negotiation.

\begin{figure*}[h!]
\centering
\includegraphics[width=\linewidth]{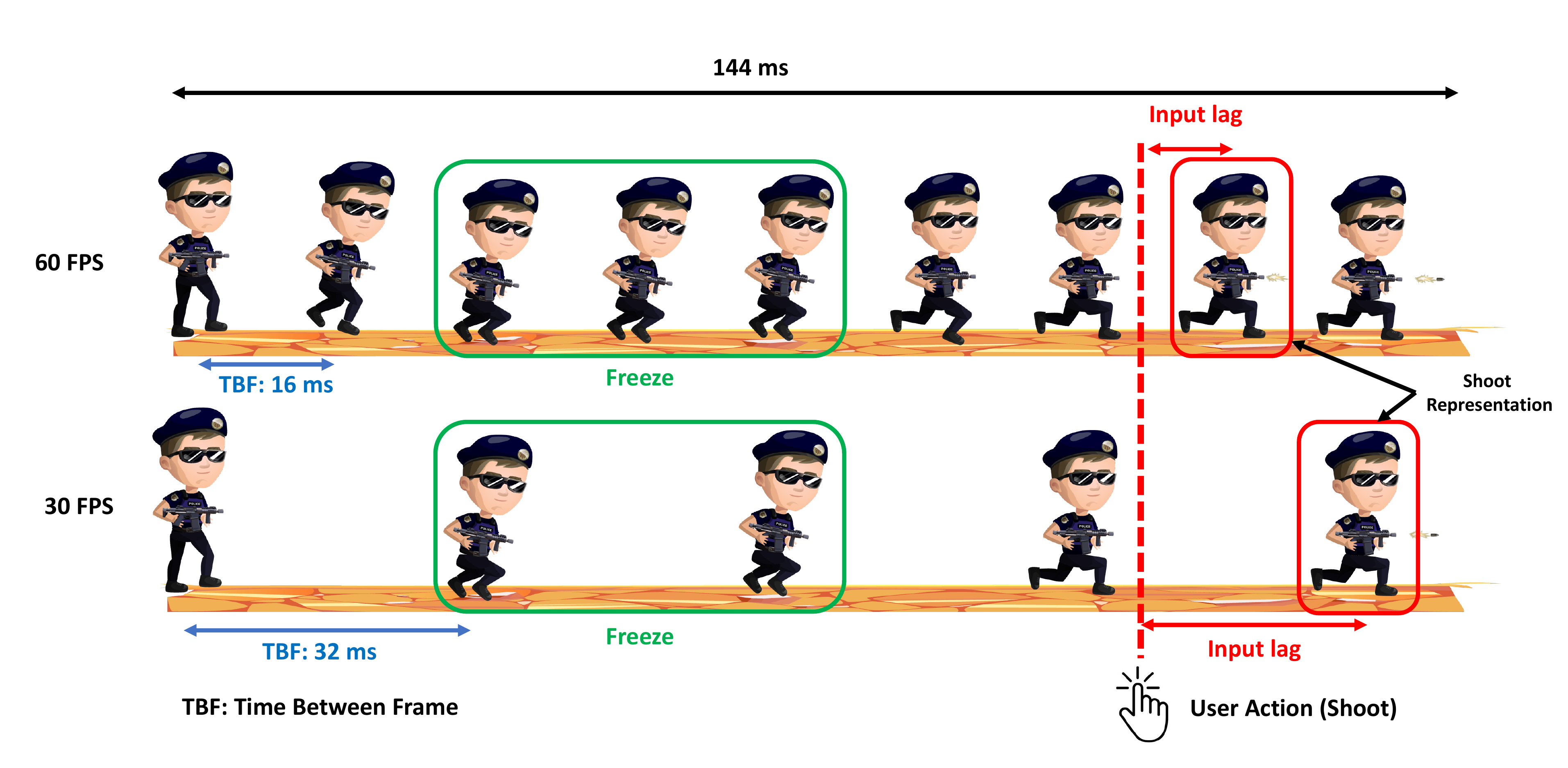}
\vspace*{-\baselineskip}
\caption{Quality indicators in an image sequence.}%\vspace{-1\baselineskip}
\label{fig:smooth}

\end{figure*}

Nevertheless, to the best of our knowledge, there are no works in the literature that help the provision of CG services based on the estimation of some parameters related to their QoE.
In this scope, the present work presents a framework that allows obtaining and estimating different KQIs of CG sessions.

Firstly, a framework for the extraction of KQI from CG services is presented.
Here, in comparison with the framework proposed in \cite{penaherrera2021measuring}, in which metrics are gathered from the CG client, metrics are calculated from user's screen, which allows getting higher layer metrics, and therefore, a closer overview of the quality perceived by the user.

Then, with the prediction of these metrics set as a goal, different Machine Learning techniques (e.g., support vector regression, random forest...) are assessed by using several predictors (i.e., the input parameters to the system) such as client configuration (e.g., resolution and fps), radio indicators (e.g., RSRP and SINR) and network configuration (e.g., resources blocks).
Furthermore, a study of the predictors' impact is carried out by the use of feature selection, which will show which predictors are the most performant for the different models to achieve a good estimation.

Hence, this study demonstrates that the performance of a CG service in terms of KQI, which are extremely difficult to know before a CG session starts, can be estimated from available easy-acquisition and heterogeneous source parameters (e.g., resolution, SINR, bandwidth) by the use of machine learning techniques.

\section{Cloud Gaming Quality Indicators}
\label{sec:cgQualityInd}

Game industry has been able to offer a new entertainment way thanks to the immersive capabilities of their services. This differentiated factor leads the QoE of gaming to be based in two basic principles: visualization and interaction.

Here, like in other network-dependent multimedia services, several Key Quality Indicators (KQIs) are considered for CG.
Thus, the following subsections give an overview of three of the main indicators to represent QoE objectively.

\subsection{Visualization}

The several challenges raised by games, as well as the events which occurs in the offered virtual world are perceived by users through film scenes. In this way, the visualization of the scenario is primordial in the gaming experience, being frequently based on image resolution, frame rate, and freeze.

Firstly, image resolution describes the granularity of the digital image. This means that a higher resolution usually leads to a more detailed scene, and therefore, a better users' perception. Nowadays, the most popular and extended resolutions are  720p (High Definition - HD), 1080p (Full High Definition - FHD), 1440p (Quad High Definition - QHD) and 4K (Ultra High Definition - UHD).

Then, the number of images per time unit (e.g., frame per second - fps) to represent the scenario sequences arises as one of the most paramount elements taken into account between gamers.
Here, Figure \ref{fig:smooth} shows the different effects that using different numbers of images for representing a scene have on the CG QoE services.

Rendering a scene using a higher number of images, which as it is observed in Figure \ref{fig:smooth} means  updating a frame in a shorter period of time, involves a more fluid and therefore, realistic movement representation, improving the user experience.
This is usually described through the frame rate metric, whose values indicate the number of frames per second (FPS) used in a video rendering.

Finally, the freeze term is used when the same frame is represented multiple times, being triggered by the lack of new frames to display.
These events usually have an extremely negative impact on the gaming experience, since it gives the feeling of image frozen which breaks the smooth movement. Besides, their frequent appearance often annoys users, who might not have control of essential events in the environment due to this freeze and therefore, lose the game.

Traditionally, all these indicators usually depend on the CPU (Central Processing Unit) and GPU (Graphics Processing Unit) capabilities of  user's devices, which are in charge of rendering the virtual environment. Nonetheless in CG, with the shift of all these tasks to the cloud, the network link hoards the prime role in the values of these indicators, since all the generated content visualized by the gamers is sent through the network.

\subsection{Interaction}
\vspace*{-0.3\baselineskip}
The main pillar of games lies in the attractive and addictive capability given to users of cooperating with the virtual world.
This leads the responsiveness of the system, which means the time from a user sends an action to the action visualization (see Figure \ref{fig:smooth}), to become one of the most critical parameters to take into account in gaming services. This indicator is commonly known as input lag or latency.

In traditional games, input lag is basically related to the time used by the user's gadgets to render the different scenes. In addition, as it can be seen in Figure \ref{fig:smooth}, frame rate also has an impact on service responsiveness. Having a higher frame rate allows perceiving a new frame earlier, and therefore, reducing the response time of the system, being the main reason why usual gamers always seek to achieve high data rates.

Lately, with the introduction of online multiplayer which allows the interaction of different users despite their localization, the input lag is now linked to the network. In this scope, most of the system's response time is caused by sending user's metadata to a common server, which must be reached by all players sharing the virtual world.

Similarly, CG's response time is generally monopolized by the delay introduced by the network in the data trading process.
Nevertheless, the volume of exchanged data in CG scenarios is much larger than in multiplayer, putting more pressure on the network link and leading to latency values that are highly dependent on the link's capabilities.
In addition, visualization settings also get into action in the input lag of the CG systems: sending content in higher resolution or frame rate will demand more requirements from the network, which might become congested and boost the latency values of the system.

\section{Measuring Key Quality Indicators}
\label{sec:dataAcquisiton}

KQI metrics are presented as a quantitative alternative to measuring the QoE of services. Their use provides a numerical way of representing the quality perceived by users, which is usually considered subjective.

Nonetheless, obtaining these high-level metrics is usually a difficult task, mostly triggered by the limited access to the application data.

Taking all this into consideration, this section describes a framework based on Python for the extraction of different KQIs from a CG platform.

\subsection{Metrics Extraction}
%\hl{Entorno controlado: 5 Acciones, escenas de 1 min, captura solo una proporcion de la pantalla.}

For the extraction of the data, a controlled environment has been created over Moonlight Client \cite{moonlight}. This platform is an open-source implementation of NVIDIA's GameStream protocol. Besides, it can stream up to 4K resolution at 120 fps over RTP protocol (Real-time Transport Protocol). 

In this sense, the controlled environment consists of a one-minute League of Legends play in which the user's screen is captured by a Python script at 144 FPS. Besides, each captured frame is marked with a time stamp.

Along this minute, 5 actions are automatically sent through the automatic action tool presented in \cite{penaherrera2021measuring}, which allows replicating the actions from the user. 
In addition, visualization settings with which the game is streamed (i.e., resolution and frame rate) are set at the beginning of each session.

Finally, once the session has ended, the images are saved and processed to obtain some high-layer quality indicators such as effective frame rate, freeze occurrences, or the system's input lag.

\subsubsection{Effective Frame Rate}

Although the server renders and sends the scene in the frame rate set at the session's beginning, their transmission through the network may differ from the frame rate visualized in the user's device. In this work, it is referred to as effective frame rate (EFPS).
In order to measure EFPS, the one-minute session is analysed frame by frame, deleting those with no difference from their predecessors.
Then, once the entire session has been swept, the total number of frames is divided by the session time, obtaining the frame rate with which the game has been shown on the user's device.

\subsubsection{Freeze occurrences}

Like in the case of effective frame rate, the fact of sending already rendered content through the network might disturb the scene's smoothness. Some events such as packet loss or jitter can hinder the representation of the whole frames of the scene, triggering freeze occurrences. 

In this scope, the calculation of freeze occurrences is highly similar to the effective frame rate procedure: each frame of the scene is analysed, identifying those consecutive which are identical. Then, thanks to each frame's timestamp, the time along the same frame that has been displayed in the user's equipment is easily obtained.

\begin{figure}[t]
\centering
\includegraphics[width=0.95 \columnwidth]{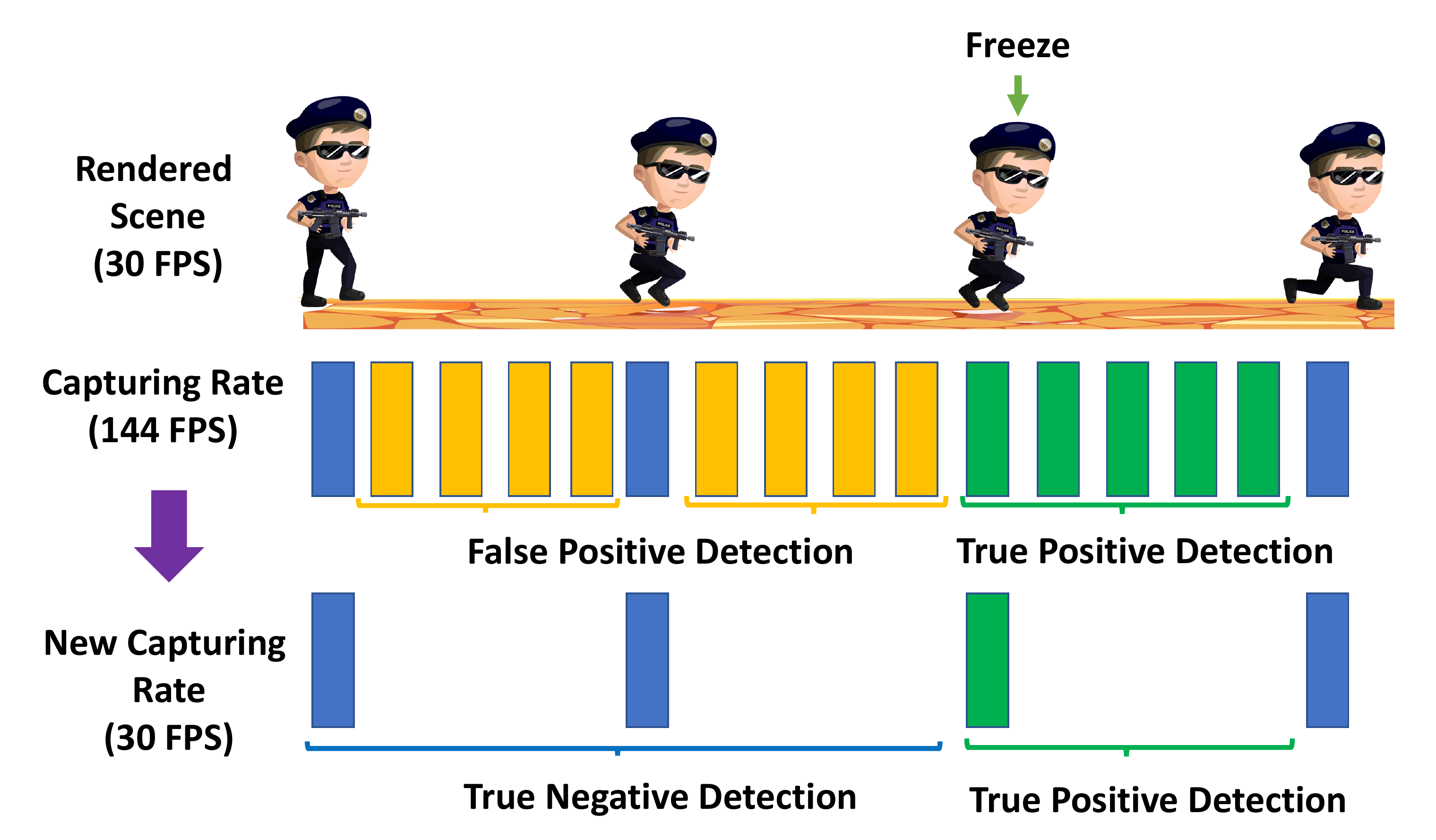}
\caption{Decimation process}%\vspace{-1\baselineskip}
\label{fig:decimation}
\vspace*{-\baselineskip}
\end{figure}

Nevertheless, the capturing rate plays a paramount role in the detection of freeze occurrences. As it is depicted in Figure \ref{fig:decimation}, a higher capturing rate than the rendered scene rate makes that some frames are the same between them, leading the algorithm to trigger false freeze (false positive detection). 

To avoid this, a decimation process is required, which means discarding frames until the capturing frame rate fits with the rendered one (Figure \ref{fig:decimation}).
In this way, through the adjustment of the capturing frame rate, the algorithm will be able to correctly detect the freeze that occurred in the scene (true negative and positive detection).

\subsubsection{Input Lag}

To obtain the system's response time, both user's action and its representation must be taken into account.
For the first one, actions' timestamps are gathered from the automatic action tool \cite{penaherrera2021measuring}.

For the latter one, it is taken the timestamp of those frames in which movement is detected. 
Here, the session is split into 5 subsets according to the timestamp of each action. 
Lately, all frames of the group are analysed one by one, finding those which differ by a 25\% of their predecessor.
This percentage of differentiation is chosen to avoid the detection of animations, a common video effect introduced in games to keep away from the feeling of freeze.

In this way, it is possible to correctly detect the exact frame and the time in which each action is represented and therefore, obtain the absolute system's response time through the difference between both timestamps.

\subsection{Dataset building}

Taking into account this approach, a dataset has been built \cite{BaenaDataset2022}.
In order to also have fully control of the network, a network-in-a-box solution has been used for the network deployment. 
This kind of solution takes advantage of GPPP (\textit{General Purpose Processing Platforms}) and SDR (Software Defined Radio) platforms to virtually deploy all entities: cellular radio and core networks. Additionally, it provides a wide range of available configurations of the network elements. 

In this way, through the framework presented in \cite{frameworkCrowd}, the network has been configured with 4 radio bandwidths (5, 10, 15, and 20 MHz), as well as different radio conditions (e.g., different power transmissions or interference), obtaining up to 16 network scenarios. 
At the same time, on these configurations, the CG service has been launched with 4 resolutions (720p, 1080p, 1440p, and 4K) and 3 different frame rates values (30, 60, and 120 fps), from which the effective frame rate, freeze occurrence and input lag of each section have been gathered.

Here, Table \ref{tab:datasetSummary} shows a description of the 16 parameters that compose the dataset as well as their source. Besides, it can be seen the minimum, maximum and average values together with the standard deviation that each parameter takes along the 3840 samples that make up the dataset.

% \usepackage{multirow}
% \usepackage{multirow}
\begin{table*}[]
\centering
\caption{\label{tab:datasetSummary}Dataset summary}
 \begin{adjustbox}{max width=\textwidth}
\begin{tabular}{@{}cccccccc@{}}
\midrule
\textbf{Source}                                             & \textbf{Indicator}   & \textbf{Unit} & \textbf{Min} & \textbf{Mean}             & \textbf{Max} & \textbf{Std.Variation} & \textbf{Description}                                 \\ \midrule
\multicolumn{1}{c|}{\multirow{3}{*}{\parbox{1.8cm}{\centering\textbf{CG Session's} \\ \textbf{  Quality}}}}   & CGlatency & ms            & 30.59        & 87.43                     & 498.65       & 36.35                  & 50th-tile of the input lag of the whole session.    \\
\multicolumn{1}{c|}{}                                       & FreezePercent        & \%            & 0            & 8.7                       & 100          & 17.6                   & Percentage of the session with a frozen image.      \\
\multicolumn{1}{c|}{}                                       & EFPS                 & fps           & 0.1          & 57.09                     & 116.17       & 29.75                  & User's perceived frame rate.                        \\ \midrule
\multicolumn{1}{c|}{\multirow{2}{*}{\textbf{CG Server}}}    & Resolution           & -             & 720p         & -                         & 4K           & -                      & Resolution set in the server for the CG session     \\
\multicolumn{1}{c|}{}                                       & fps                  & fps           & 30           & -                         & 120          & -                      & Frame rate set in the server for the CG session     \\ \midrule
\multicolumn{1}{c|}{\multirow{7}{*}{\textbf{UE}}}           & PING\_avg            & ms            & 1            & \multicolumn{1}{l}{74.88} & 895          & 84.63                  & Round-trip-time between UE and Server during the session               \\
\multicolumn{1}{c|}{}                                       & PING\_Radio\_Loss\%   & \%            & 0            & 0.23                      & 25           & 1.27                   & Ping percentage loss in the radio part.  \\
\multicolumn{1}{c|}{}                                       & PING\_Host\_Loss\%    & \%            & 0            & 0.61                      & 25           & 2.02                   & Ping percentage loss in the whole path.   \\
\multicolumn{1}{c|}{}                                       & RSRP                 & dBm           & -104         & -71                       & -50          & 13.06                  & Avg. received power from the reference signal. \\
\multicolumn{1}{c|}{}                                       & RSRQ                 & dB            & -8           & -4.1                      & -3           & 1.1                    & Quality of the received reference signal.           \\
\multicolumn{1}{c|}{}                                       & RSSI                 & dB            & -95          & -56.51                    & -51          & 8.5342                 & Strength of the received radio signal.              \\
\multicolumn{1}{c|}{}                                       & SINR                 & dBm           & 6            & 17.52                     & 26           & 6.85                   & Signal-to-interference-plus-noise ratio.            \\ \midrule
\multicolumn{1}{c|}{\multirow{4}{*}{\parbox{1.2cm}{\centering\textbf{Base} \\ \textbf{Station}}}}  & n\_rb\_dl            & RB            & 25           & -                         & 100          & -                      & No. available resource blocks (RB) to assign.       \\
\multicolumn{1}{c|}{}                                       & cqi                  & -             & 0            & 12                        & 15           & 2.53                   & Channel quality indication reported.                \\
\multicolumn{1}{c|}{}                                       & pucch\_snr           & dBm           & -11.39       & 14.66                     & 44.14        & 15.92                  & PUCCH SNR reported to the Base Station.             \\
\multicolumn{1}{c|}{}                                       & pusch\_snr           & dBm           & -25.78       & 15.57                     & 36.65        & 8.67                   & PUSCH SNR reported to the Base Station.             \\ \midrule
\end{tabular}
\end{adjustbox}

\end{table*}

\section{Estimation of Key Quality Indicators}
\label{sec:estimationMethod}
As it has been seen in the previous section, measuring CG KQI entails several processes which might take long time periods. Besides, some tasks might demand a high amount of computational resources to be carried out (e.g., screen capturing).

\begin{figure}[h]
\centering
\includegraphics[width= \columnwidth]{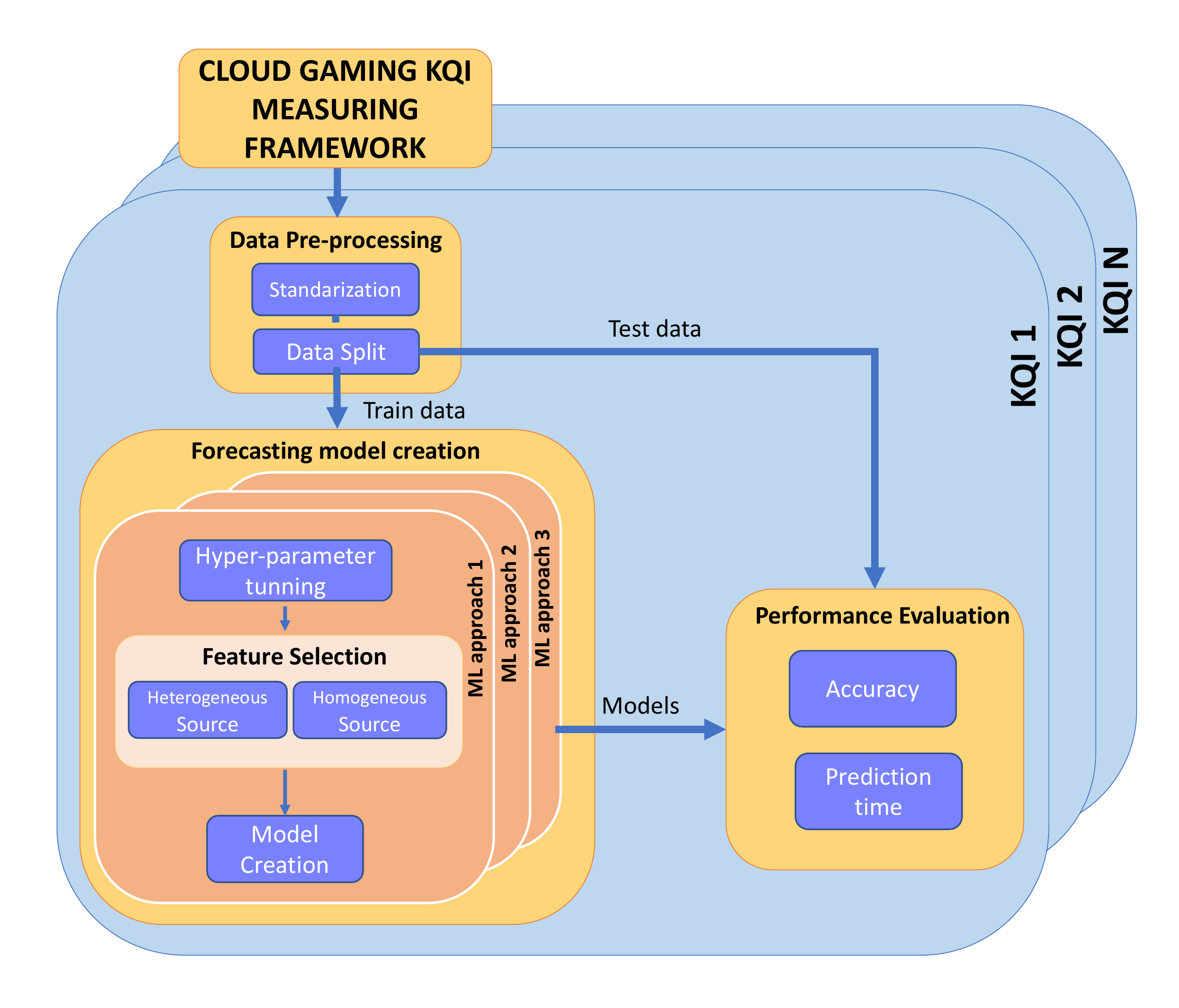}
\caption{KQI estimation flowchart}%\vspace{-1\baselineskip}
\label{fig:trainFramework}
\vspace*{-\baselineskip}
\end{figure}

In order to avoid that, this section provides a machine learning approach to estimate such KQI from network parameters.
The methodology followed to their prediction is depicted in Figure \ref{fig:trainFramework}. 

This approach aims to search the best models to reach a good estimation of these KQIs.
To do so, firstly, a pre-processing stage is performed over the data, in which values are standardised and split into train and test subsets.

Based on the training subset, several models following different machine learning approaches are created for the prediction of each KQI (i.e., \textit{CGlatency}, \textit{FreezePercent}, \textit{EFPS}). 
To do so, by grid searching it is obtained the configuration (hyperparameters) with which the models fit the most with the data to predict. 

With these configurations, the different techniques are trained considering  both all the input features of the dataset and a meaningful subset of them by the use of Feature Selection (FS).   
This selection is carried out following the highest scores with which the different features are weighted. Additionally, it is taken into account the source of the different features (e.g., UE or BS), allowing the assessment of predictions from only one element.

Finally, once all the models are trained, their performance in terms of accuracy and execution time is evaluated using the test subset.
The following subsections provide a  more detailed description of each stage.

\subsection{Data Pre-processing}

As it is seen in Table \ref{tab:datasetSummary}, the different indicators come from heterogeneous sources (e.g., UE and Base Station), as well as show different ranges (e.g., \textit{SINR} from UE shows values from 6 to 26, meanwhile \textit{pusch\_snr} takes values between -25 and 36) and units (e.g., \textit{PING\_avg} and  \textit{RSRP} are measured in ms and dBm respectively).

In this scope, the application of standardization techniques becomes to play an important role, since they allow overawing on the dataset to bring all the inputs on the same scale. This makes easier the analysis of the indicators for the creation of the models, and therefore, it leads to faster convergence of the different ML algorithms.

Accordingly, in this work the data is standardised following a Min-Max Scaler. This method regularises the data to unit variance as follows:

\iffalse
\hl{Update to Min-Max Scaler}
This method removes the mean and scales the data to unit variance. To do so, the value of the different parameters (denoted as $x_{norm}$) is determined as  
\fi
\begin{equation}
    x_{i-scaled} = \frac{x_{i}-x_{min} }{x_{max} - x_{min}},
\end{equation}

being $x_{i}$ the original value of the parameter, $x_{max}$ the maximum value along with the whole set of the parameter, and $x_{min}$ the minimum one.

Additionally, the dataset is randomly split into training and test subsets, containing 70\% and 30\% of the samples respectively.

\subsection{Estimation model building}
\label{sec:estimationModels}
The creation of the models is the paramount stage to performing a high-accuracy estimation. 
Hence, this section provides an overview of the techniques taken into account, as well as the steps carried out to create different forecast models.

\subsubsection{Machine Learning techniques}

Since this approach pursues to estimate high-level layer metrics (i.e., KQI), which are difficult to obtain in a non-controlled situation, from others available which are easy to acquire in any situation, supervision learning (SL) techniques are considered.

SL approaches map one or several inputs with an output by analysing a set of examples. These examples, with which the different techniques are trained, are known as labeled data.

In this scope, five of the most extended SL techniques are considered in this approach:

\begin{itemize}
  \item Linear Regression (LR) is a mathematical procedure that is commonly used to approximate the relationship between one dependent variable and one or more independent variables. To do so, this procedure follows to find the linear combination of the features which minimises the residual sum of squares between the real and predicted values.
  
  \item K-Neighbors Regression (KNR) is a regression based on the K-nearest neighbors algorithm (k-NN). This algorithm uses feature similarity to predict values. 
  This means that the prediction is assigned to a point based on how closely it resembles the k-points (or neighbors) in the training set.
  Finally, the value assigned to the prediction is performed by averaging (or weighting) the different neighbors. 
  
  \item Support Vector Regression (SVR) follows to get a hyperplane that fits with most of the data samples. To do so, the data is transformed  through a set of mathematical functions (also called \textit{kernel}) to a higher-dimensional space, which eases the definition and searches for an optimal regression hyperplane.
  At the same time, based on a tolerance parameter $\epsilon$, boundary lines are defined around the hyperplane with the aim of neglecting any fluctuation within these margins. In this way, unlike other regression procedures, SVR attempts to minimise the generalised error within a range.
  
  \item Kernel Ridge (KRR) is a regression approach like SVR, that uses the kernel trick to project the data in other-dimensional spaces and ease the quest for a regression function. However, the main disparity between SVR and KRR resides on the loss function. Here, like in LR, this technique follows to minimise the squared error loss to find the optimal model but using a regularisation parameter $\alpha$ to shrink the estimation variance.

  \item Random Forest (RF) consists of a large number of decision trees whose outputs are averaged to get the predicted values.
  Such decision tree models are created through bootstrap aggregation (bagging), which consists of random subsets with the replacement of the training data.
  This process eases the creation of uncorrelated models, which leads to overcoming the decision tree's sensitiveness with the trained data. In this way, their assembly operation allows obtaining robust regression models, since trees will protect each other from individual errors, boosting the performance of the model.

  \item Artificial Neuronal Network (ANN) is a machine learning technique based on the architecture of the human brain.                                         
  Thus, they are conformed by several nodes, also called artificial neurons, which are distributed along with different layers (input, hidden, and output layers). 
  Such artificial neurons are in charge of computing and weighting the data by activation functions and sending their output to the next layer of the network. Hence, this approach achieves the creation of a complex regression model.

 \end{itemize}

\subsubsection{Hyper-parameter tunning}

Machine learning algorithms are open to multiple designs through different \textit{hyper-parameters}. 
Their setting highly depends on the nature of the problem and the data, so an unappropriated configuration of them may lead to debase the model prediction.
Therefore, it is primordial to carry out a hyper-parameter tune process that allows searching their optimal settings.

In this scope, an exhaustively searching of the most important hyper-parameters has been separately made for the different KQIs that we follow to predict (i.e., \textit{CGlatency}, \textit{FreezePercent} and \textit{EFPS}).

Table \ref{tab:hyperparms} provides a brief description of the different hyper-parameters tuned in this work by grid searching. Here, the hyper-parameters fixed are those which allow minimising the Mean Absolute Error (MAE) of the training dataset. This figure of merit is defined as below:

\iffalse
\begin{equation}
    R^{2} = \frac{\sum(\widehat{y} - \overline{y})^2}{\sum(y - \overline{y})^2}
\end{equation}
where $\widehat{y}$ is the predicted value, and $y$ and $\overline{y}$ are the original value and their mean, respectively.

\fi

\begin{equation}
    MAE = \frac{1}{N}\sum_{i=1}^{N}|y_{i} - \hat{y_{i}}|
\end{equation}

where $\widehat{y_{i}}$ represents the predicted value, and $y_{i}$ is the original one.

\subsubsection{Feature Selection}

Generally in the machine learning scope, the higher input features of a model the better estimation accuracy it provides. 
Nonetheless, the use of a large number of predictors leads to elevated computational cost and time-use pre-processing and training process. Consequently, it is required a trade-off between accuracy and the number of predictors.

On its behalf, the use of features from different sources might enrich the performance of the estimation.
In this work, the integration of the KQI measuring framework with a controlling network environment eases the gathering of features from different elements of the architecture (i.e., UE, Base station).

However, in a non-controlling scenario, obtaining all these features might become tricky. 
On the one hand, the gathering of heterogeneous-source metrics is not trivial, since it is necessary to carefully manage the acquisition of them in the different elements for their merge based on timestamp. 
Besides, this process might lead that not all indicators will be available at the same time, and therefore, introducing a delay in the prediction.
On the other hand, network metrics are usually safeguarded at the operator level. This fact difficulty their availability and leads their provision to external parties being based on the network's high-level information along with a time slot. Moreover, the high computational cost which supposes the recollection of a large number of metrics drives operators to minimise the parameters to save.

In this scope, a study of the effect of the use of different predictors is carried out through the selection of diverse features. 
Thus, different models are created taking into account several feature subsets of the whole dataset.
To do so, an automatic feature selection is applied following two criteria: features from all the dataset or features from one source together with CG stream configuration. 

%In order to carry out an automatic selection, features are taken following the k-highest scores which are assigned to each one. These scores are assigned by the analysis of dependency between the input features and the target one.

%In this sense, mutual information (MI) has been proposed in this work as a score function, which provides the amount of information that one random variable possesses about another one.

\begin{figure}[h]
\centering
\subfigure[Cloud Gaming latency]
{\label{fig:MIlatency}\includegraphics[width=0.45\textwidth]{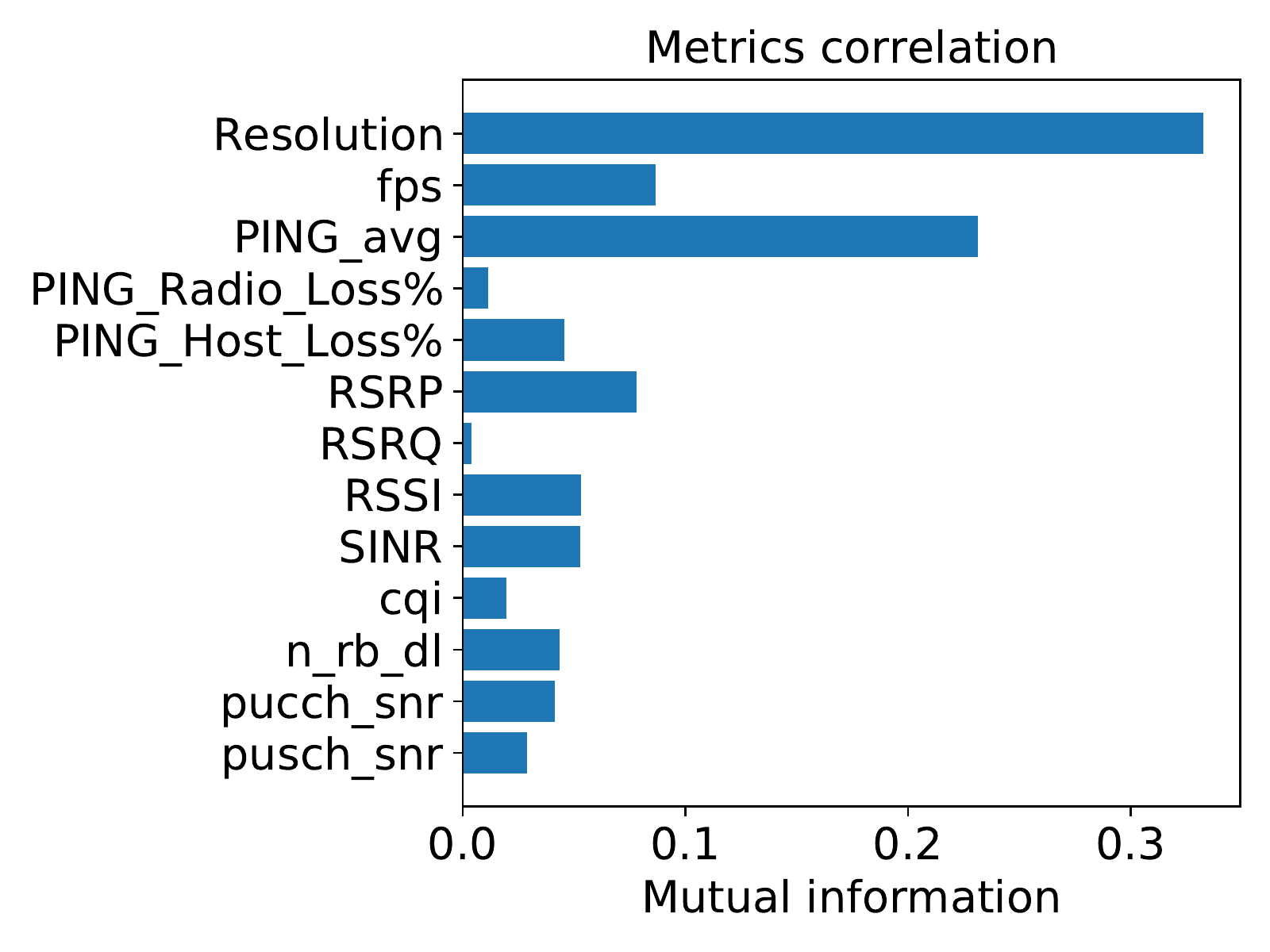}}
\subfigure[Freeze Percent]{\label{fig:MIFP}\includegraphics[width=0.45\textwidth]{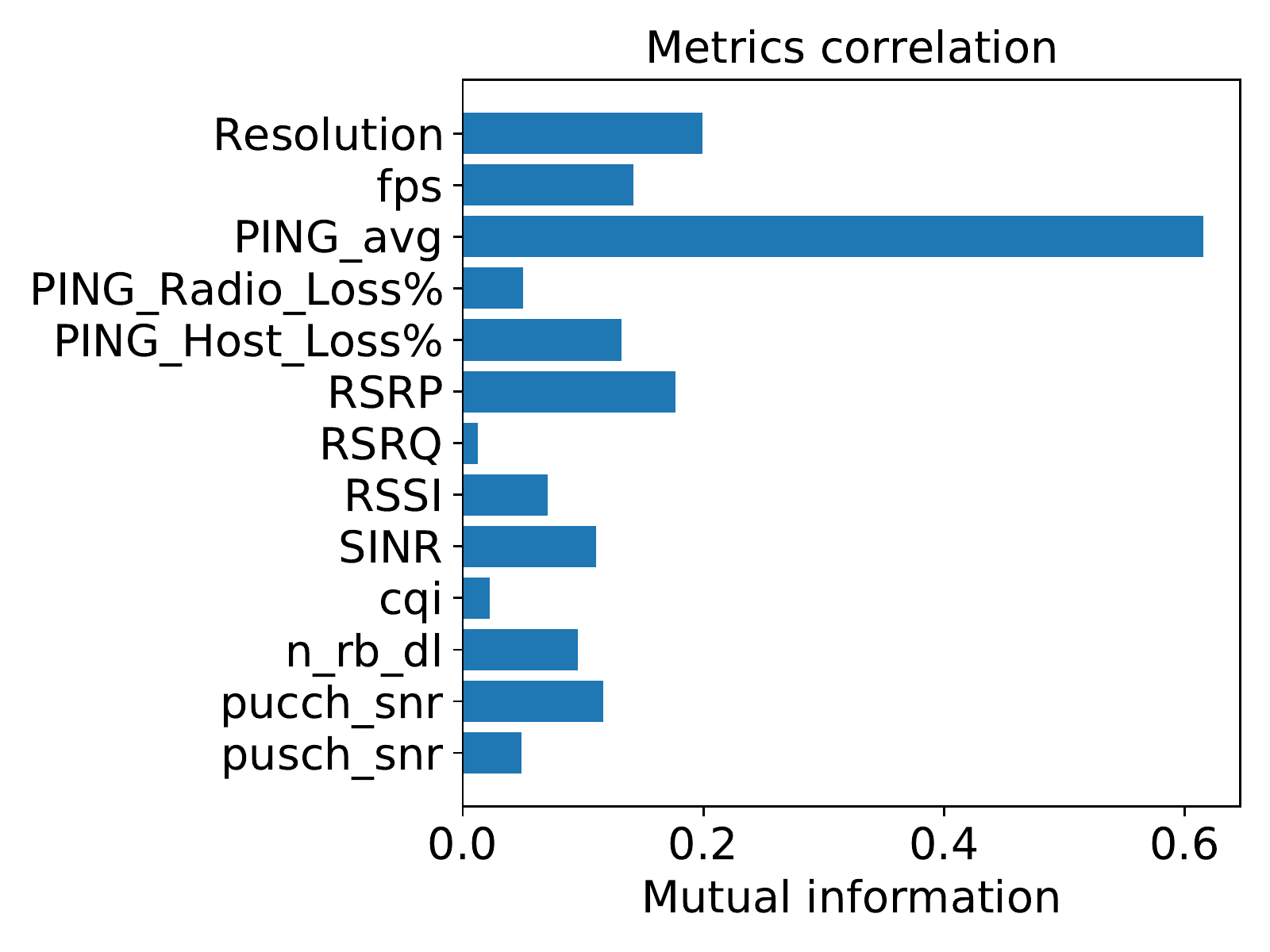}}
\subfigure[EFPS]{\label{fig:MIefps}\includegraphics[width=0.5\textwidth]{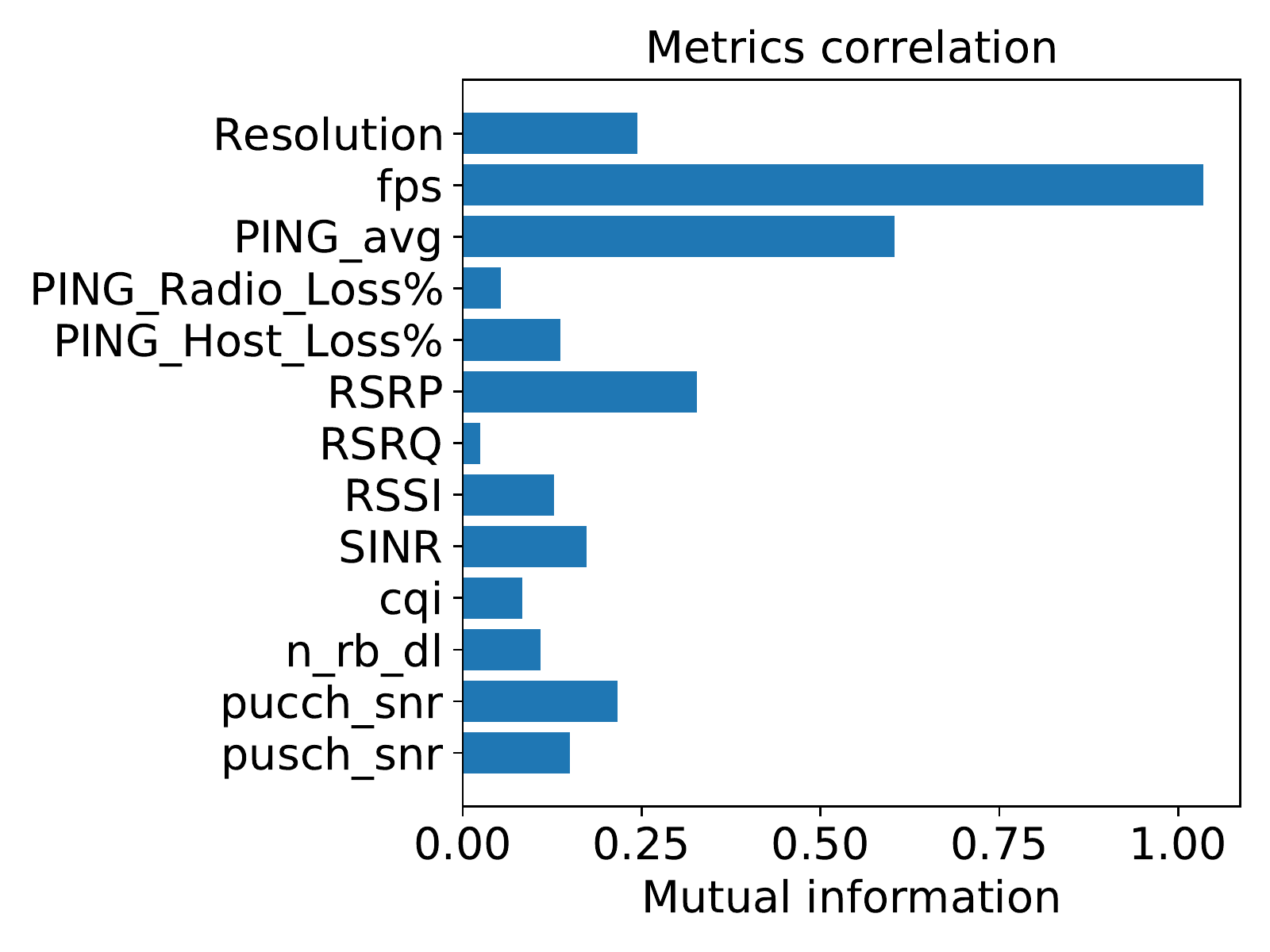}}
\caption{Metric correlation through mutual information}
\label{fig:mutualInformation}
\end{figure}

In order to carry out an automatic selection, features are taken following the k-highest scores which are assigned to each one. These scores are assigned by the analysis of dependency between the input features and the target one, usually being set by Pearson's correlation. However, mutual information (MI) has been proposed in this work as a score function, which provides the certainty level between two variables as follows:

\begin{equation}
    I(X,Y) = \sum_{i=1}^{n}\sum_{j=1}^{m} P({x_{i}}, {y_{j}}) \cdot log\frac{P({x_{i}}, {y_{j}})}{P({x_{i}}) \cdot P({y_{j}})}
\end{equation}

In this sense, MI defines the scores based on relationships beyond the linear ones obtained with Pearson's correlation, such as non-monotonic relationships.
Figure \ref{fig:mutualInformation} shows the mutual information between the different input features of the dataset and the KQIs to predict.

\iffalse

\begin{figure}[h]
\centering
\includegraphics[width= \columnwidth]{images/metrics_correlation.pdf}
\caption{Metric correlation through mutual information}%\vspace{-1\baselineskip}
\label{fig:mutualInformation}
%\vspace*{-\baselineskip}
\end{figure} 
\fi

%----------------------------------------------------------------------

Here, although the values mainly depend on the output,
it is generally seen how the configured resolution, the frame rate of the session, and the ping between the client and the server are positioned as the most influential features for each KQI.

Besides, other parameters such as RSSI, SINR, and pucch\_snr seem to have the same influence over the different KQIs.
Conversely, RSRQ and CQI seem to have a very slight dependence on the target parameters.

In this way, the automatic selection will take the k-highest influential features for each KQI. 
On its behalf, the feature's source-based selection 
takes into consideration the element from which the features are generated.
For instance, a model can be created to estimate a KQI by only taking parameters of the session configuration and those which are available in the UE (see Table \ref{tab:datasetSummary}).

In this sense, this criteria allows analysing the performance of the models if there are only indicators (or a part of them) from an element of the entire communication architecture. Here in this case, in order to also consider that not all of the entire indicators from an element may be available, MI has also been used to analyse the estimation's performance through the different number of input features.

% \usepackage{booktabs}
% \usepackage{multirow}
\begin{table*}[]
\centering
\caption{Hyper-parameters tunning summary}
\label{tab:hyperparms}
\begin{adjustbox}{max width=\textwidth}
\begin{tabular}{@{}cccccc@{}}
\toprule
                                          & \textbf{Hyperparameter}      & \textbf{Description}                                               & \textbf{Latency}   & \textbf{Freeze}    & \textbf{EFPS}              \\ \midrule
\multicolumn{1}{c|}{\multirow{4}{*}{\textbf{KNR}}} & n\_neighbors        & Number of neighbors                                       & 12        & 4         & 4                 \\
\multicolumn{1}{c|}{}                     & p                   & Power parameter (only for Minkowski metric)               & -         & -         & -                 \\
\multicolumn{1}{c|}{}                     & weight              & Prediction weight function                                & distance  & distance  & distance          \\
\multicolumn{1}{c|}{}                     & metric              & Distance metric for finding neighbors                     & Manhattan & Manhattan & Manhattan         \\ \midrule
\multicolumn{1}{c|}{\multirow{4}{*}{\textbf{SVR}}} & kernel              & Function to transform the data                            & Poly      & Linear    & RBF               \\
\multicolumn{1}{c|}{}                     & epsilon             & Tolerance parameter within which no penalty is associated & 3.5       & 0.5       & 2.5               \\
\multicolumn{1}{c|}{}                     & degree              & Degree of the polynomial function (only for poly kernel)  & 5         & -         & -                 \\
\multicolumn{1}{c|}{}                     & C                   & Regularisation parameter through a squared l2 penalty     & 10        & 10        & 300               \\ \midrule
\multicolumn{1}{c|}{\multirow{3}{*}{\textbf{KRR}}} & alpha               & Regularisation strength                                   & 1.0       & 1.0       & 1.0               \\
\multicolumn{1}{c|}{}                     & kernel              & Function to transform the data                            & Poly      & Poly      & Poly              \\
\multicolumn{1}{c|}{}                     & degree              & Degree of the polynomial function (only for poly kernel)  & 6         & 6         & 6                 \\ \midrule
\multicolumn{1}{c|}{\multirow{3}{*}{\textbf{RF}}}  & n\_estimators       & Number of trees conforming the forest                     & 70        & 80        & 90                \\
\multicolumn{1}{c|}{}                     & max\_depth          & Maximum depth of the trees                                & 10        & 20        & 20                \\
\multicolumn{1}{c|}{}                     & criterion           & Figure of merit to measure the quality of a split         & MSE       & MSE       & MAE               \\ \midrule
\multicolumn{1}{c|}{\multirow{4}{*}{\textbf{ANN}}} & No. Layers          & Number of layer conforming the Neural Network             & 4         & 3         & 6                 \\
\multicolumn{1}{c|}{}                     & No. Neurons         & Number of neurons in hidden layers                        & (20,14)   & (10,5)    & (120, 40, 20, 10) \\
\multicolumn{1}{c|}{}                     & Activation function & Function used to determine the output of each layer       & Relu      & Relu      & Tanh              \\
\multicolumn{1}{c|}{}                     & Optimiser           & Function to optimise ANN attributes (e.g., weights)        & Adam      & RMSprop   & Adam              \\ \bottomrule
\end{tabular}
\end{adjustbox}
\end{table*}

\subsection{Model assessment}
Once the models are built by using the techniques previously described, their performance is assessed through the Mean absolute scaled error (MASE) \cite{MASEref}.

This metric makes it possible to represent how good is a model estimation compared to naive forecasting. In other words, MASE provides how the use of regression models improves the estimation error obtained without establishing causal factors. It is denoted in terms of MAE as:

\begin{equation}
    MASE = \frac{MAE}{MAE_{in-sample,naive}}
\end{equation}

From the equation, it is seen that MASE returns a scaled error of the model prediction regarding a naive forecasting. 

Hence, values closer to 1 are translated that the model behaves as a naive forecasting. This can be translated into that the regression model is not introducing any enhancement. 
On its behalf, values of MASE greater than 1 indicate that the model is demoting the prediction and consequently, smaller errors can be reached through a naive estimation.  
Lastly, values below 1 suggest that models are acceptable. Their performance becomes richer while MASE values become smaller.

Therefore, like MAPE, MASE is a scale-free figure of merit which allows comparing the forecast performance  regardless of the scale of the data. 
Nevertheless, MASE offers several advantages over MAPE.

Unlike MAPE, it puts the same penalty for both positive and negative errors. 
This issue often leads to the use of sMAPE (symmetric MAPE) instead of MAPE. Nonetheless, MASE never provides infinite or undefined values even if there are zero values in the dataset, as is the case of MAPE and sMAPE.

As see in Table \ref{tab:datasetSummary}, some of the CG KQIs could present values equal or close to zero (i.e., \textit{FreezePercent} and \textit{EFPS}). For this reason, and thanks to the characteristics previously described, MASE is positioned as the most adequate metric for the evaluation of prediction accuracy.

Finally, in addition to the forecast effectiveness, the computational cost of each model is also considered in terms of prediction time.

\section{Evaluation}
\label{sec:evaluation}
In order to evaluate the estimation system, this section offers a performance analysis of the different models.
Table \ref{tab:hyperparms} provides the set of hyper-parameter with which each algorithm minimises the prediction error. 
From this point, all assessed models will follow these settings, taking into account the used algorithm and the KQI to predict.

On its behalf, as described in Section \ref{sec:estimationMethod}, each regression approach is tested with different predictors. 
Thus, models are created with predictors from all the dataset or considering their source (i.e., combination of CG server parameters with metrics from the UE or the Base Station).

The following subsections present the assessment of such models for the different metrics to estimate. 
To do so, and as aforementioned, both estimation error and computational cost will be taken into account. 

\begin{figure*}[h]
\centering
\subfigure[Cloud Gaming latency]{\label{fig:latency_mase}\includegraphics[width=0.45\textwidth]{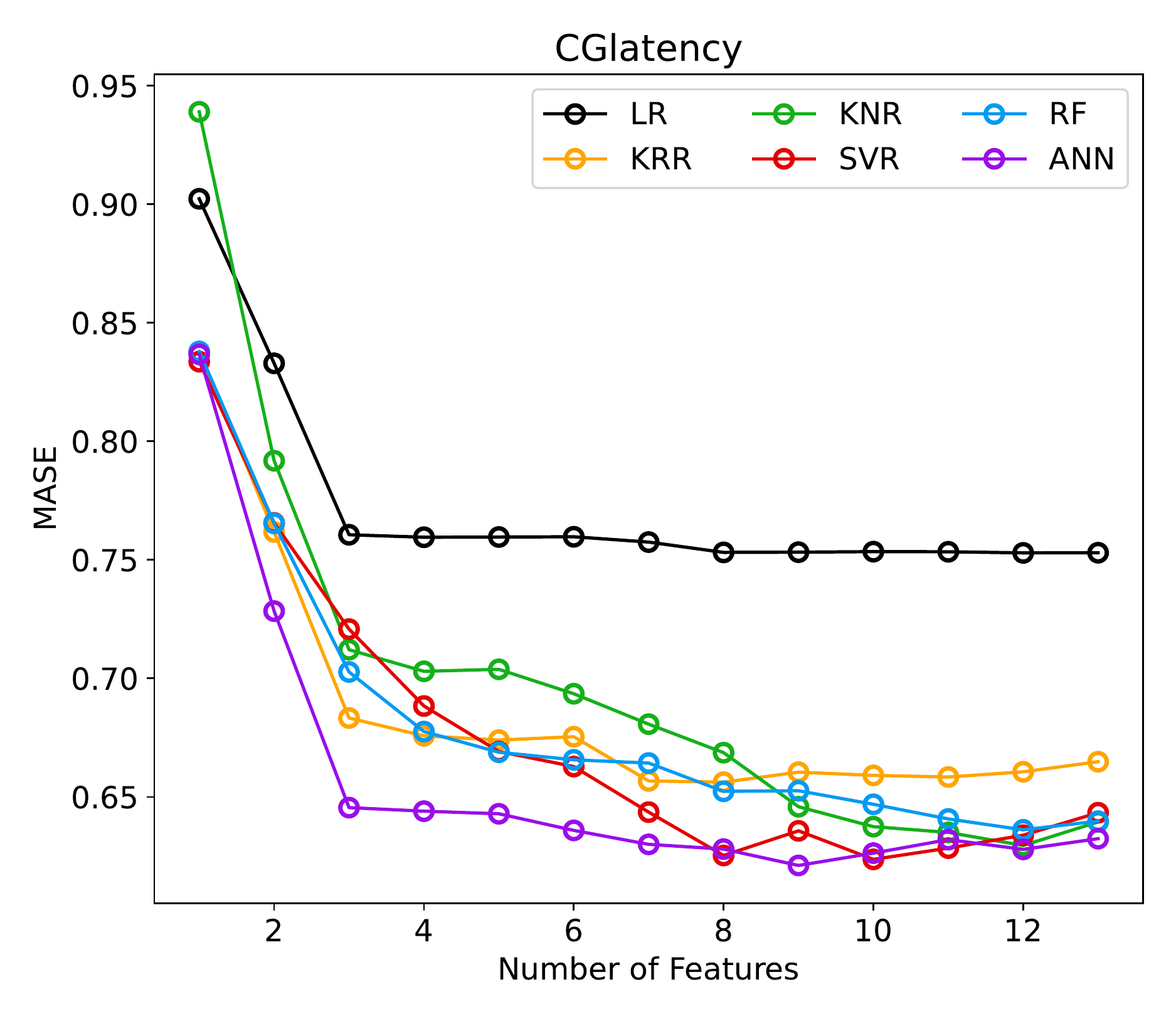}}
\subfigure[Freeze Percent]{\label{fig:freeze_mase}\includegraphics[width=0.45\textwidth]{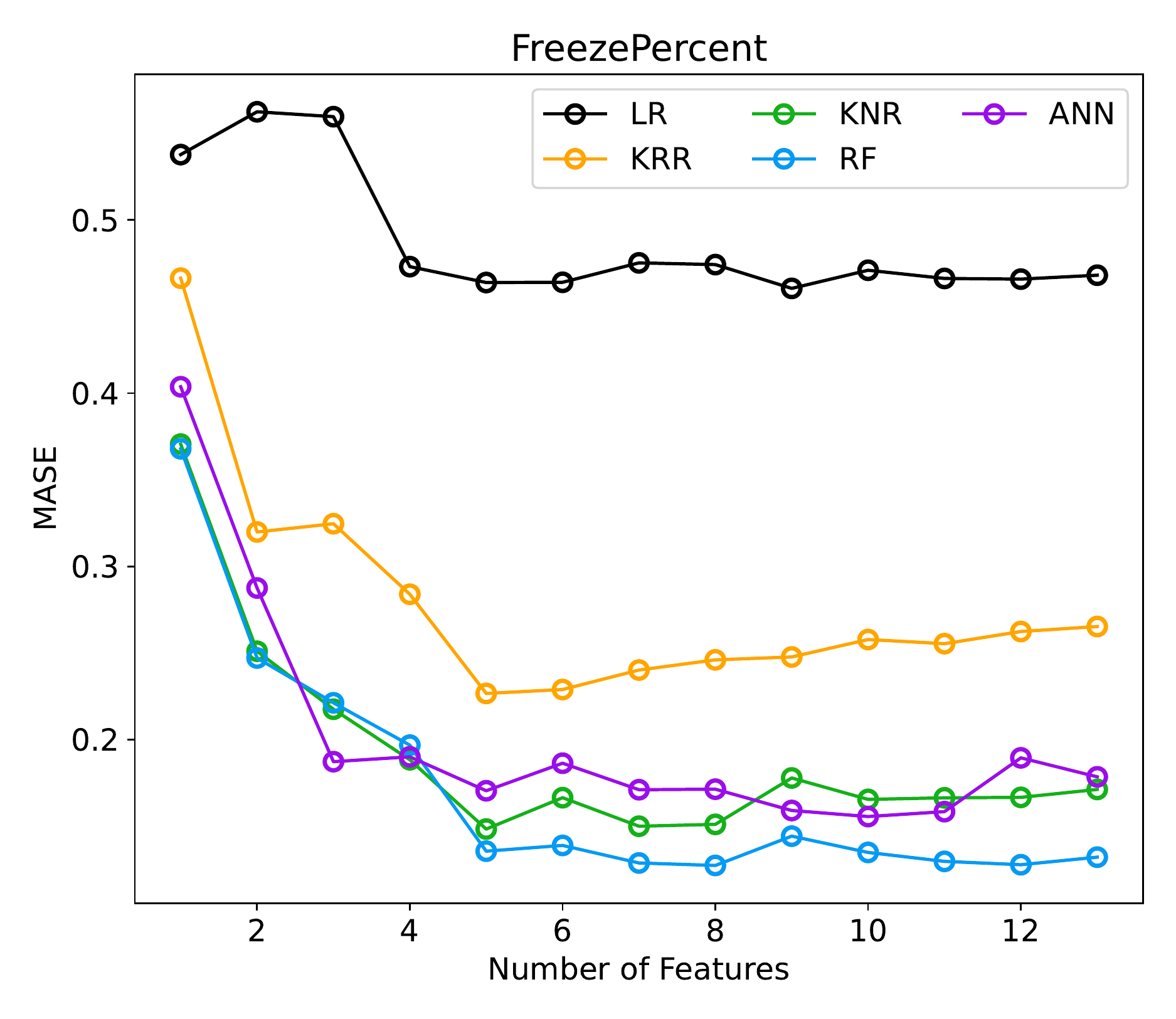}}
\subfigure[EFPS]{\label{fig:efps_mase}\includegraphics[width=0.45\textwidth]{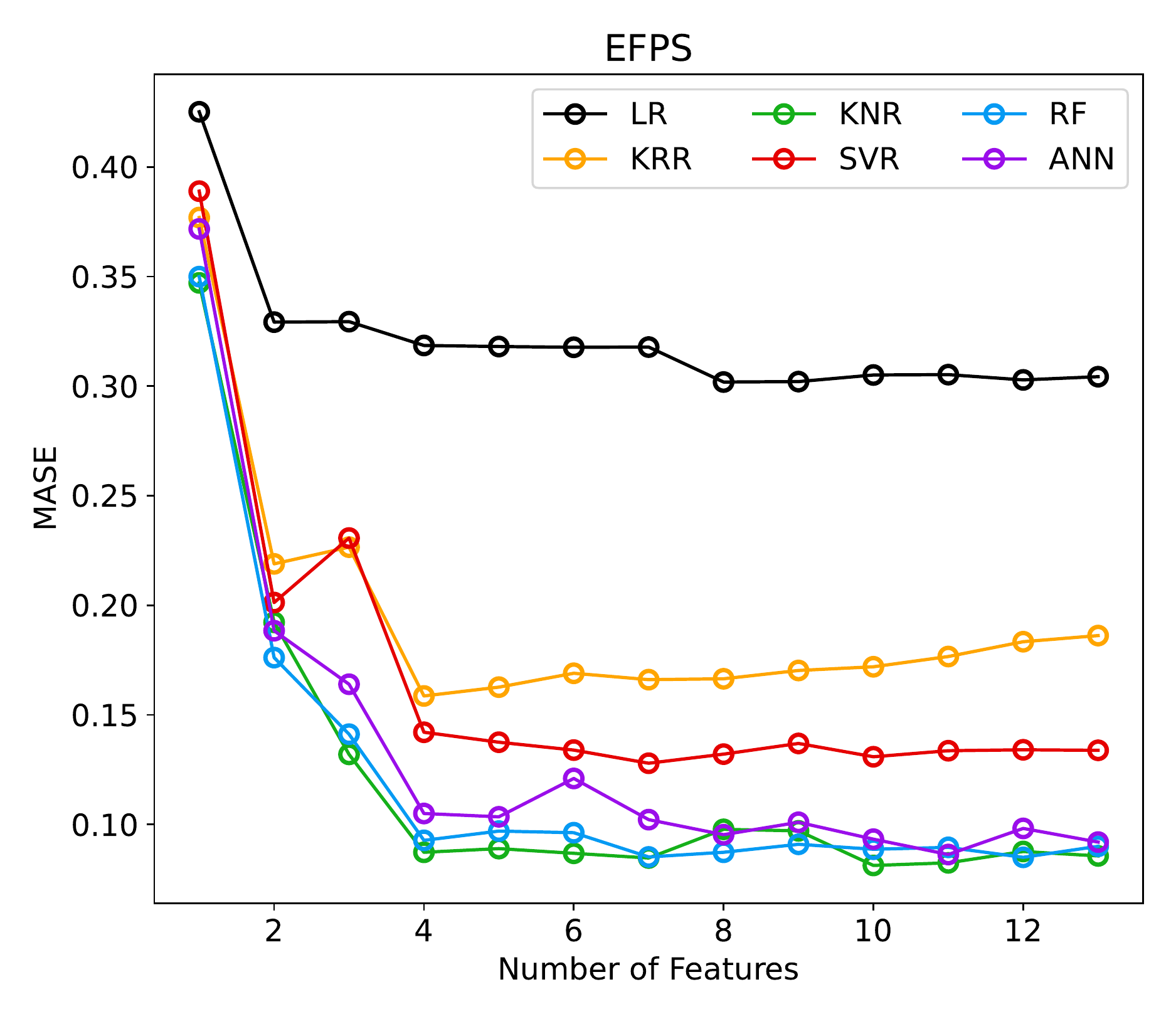}}
\caption{MASE evolution}
\label{fig:MASE_evol}
\end{figure*}

\begin{figure*}[h]
\centering
\subfigure[Cloud Gaming latency]{\label{fig:latency_mase_sources}\includegraphics[width=0.45\linewidth]{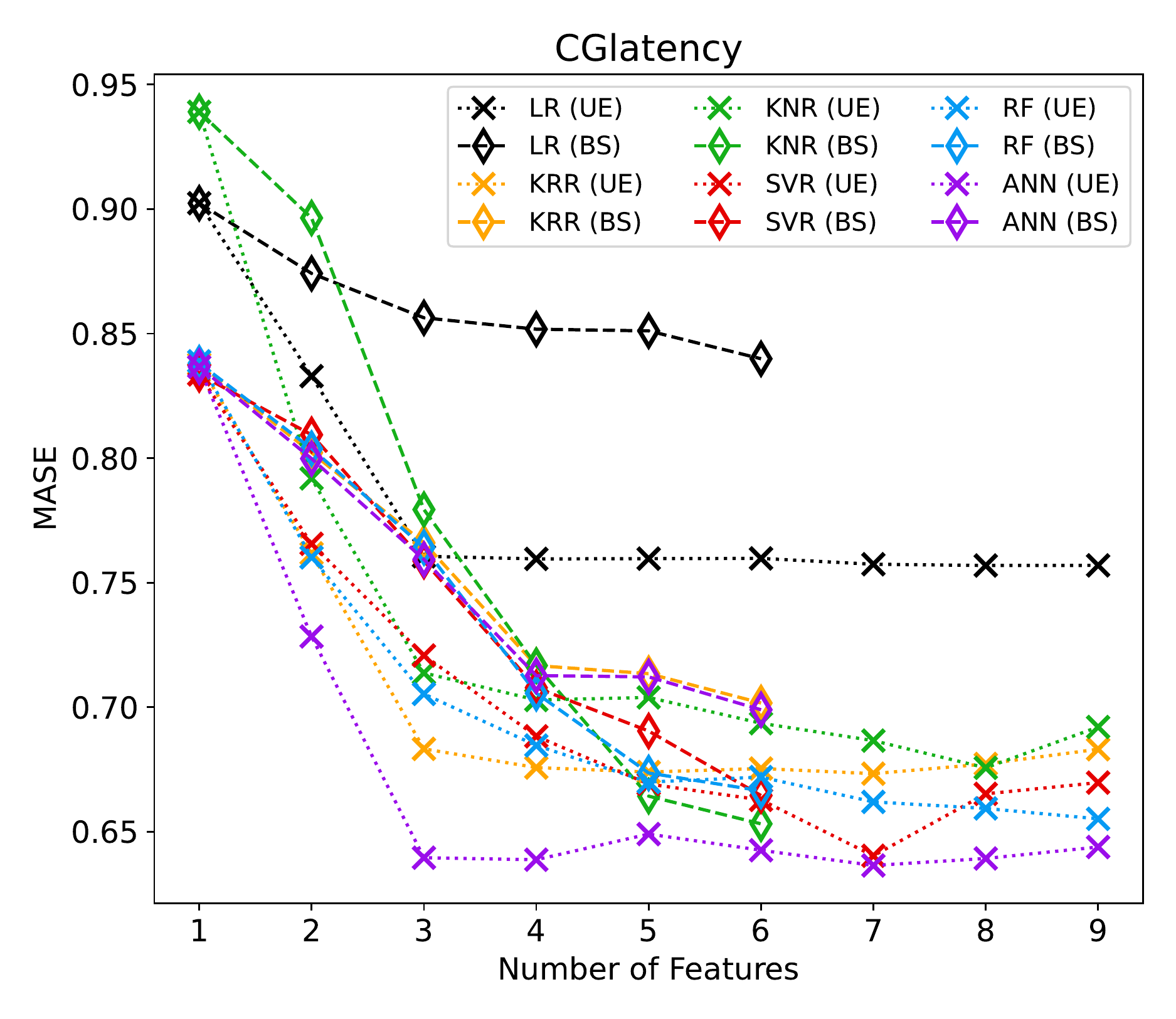}}
\subfigure[Freeze Percent]{\label{fig:freeze_mase_sources}\includegraphics[width=0.45\linewidth]{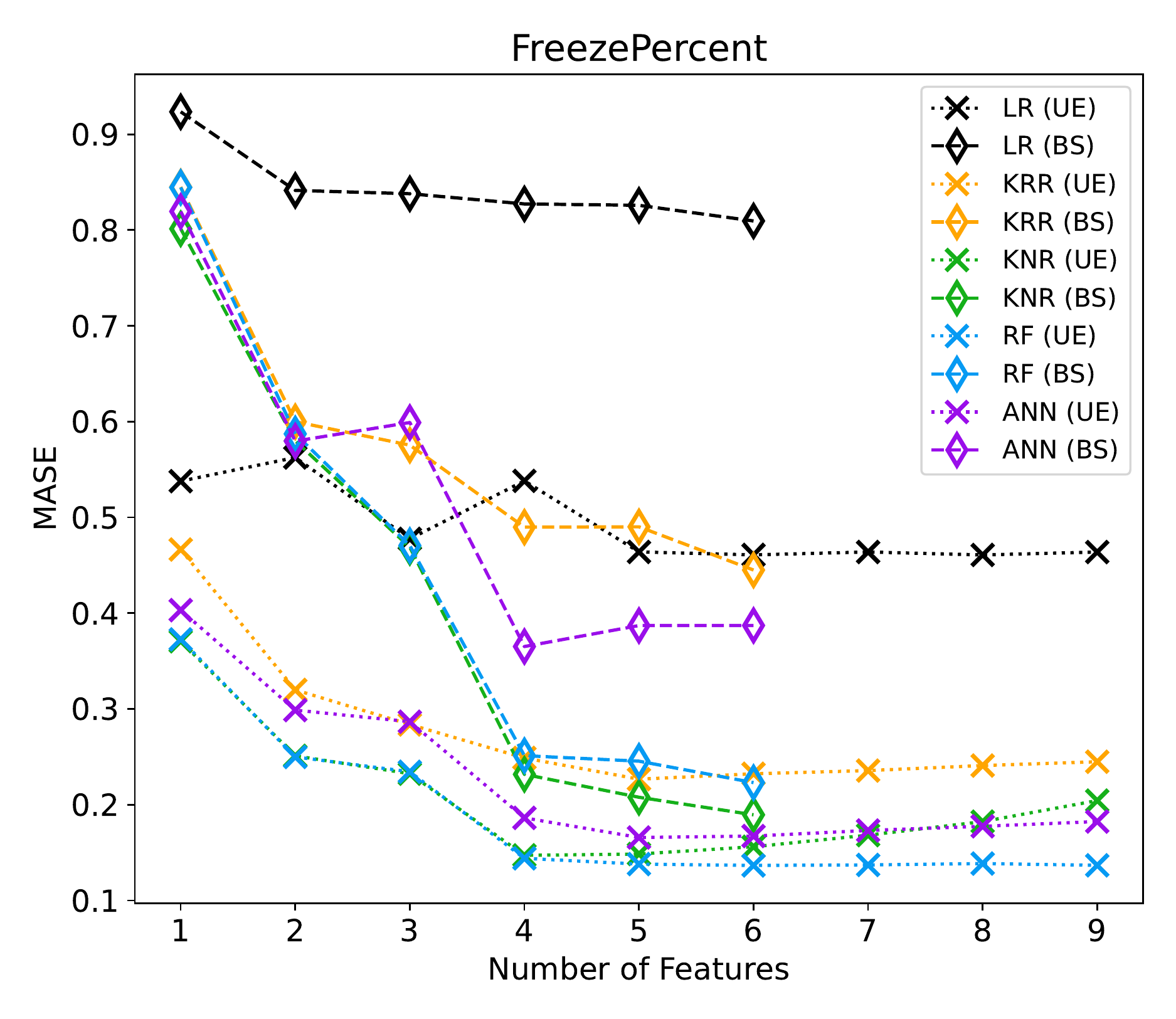}}
\subfigure[EFPS]{\label{fig:efps_mase_sources}\includegraphics[width=0.45\linewidth]{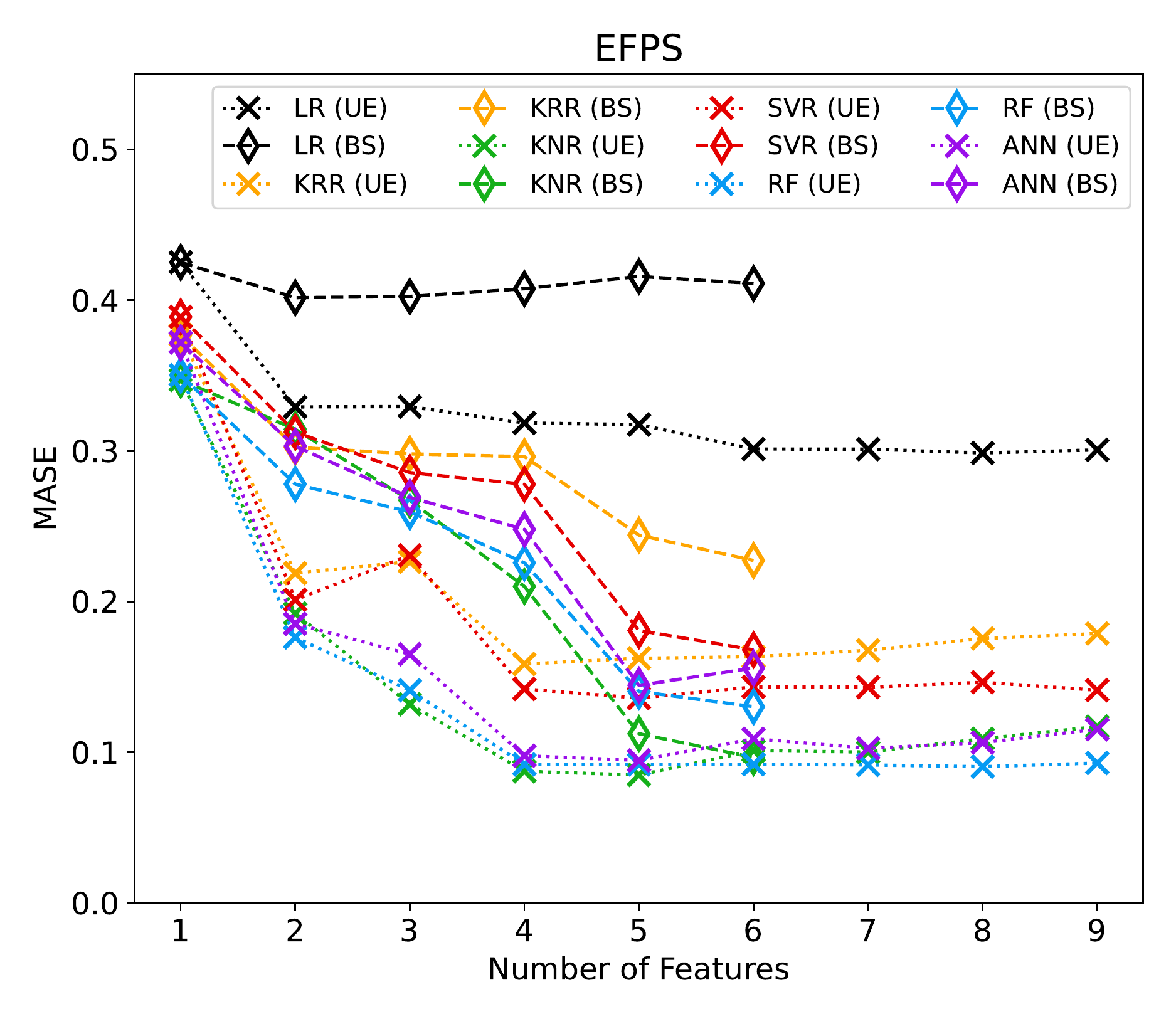}}
\caption{MASE evolution considering the source}
\label{fig:MASE_evol_sources}
\end{figure*}

\subsection{Estimation error}

Figure \ref{fig:MASE_evol} depicts the evolution of the MASE for the prediction of the different KQIs through the ML algorithms and the number of input features.
Such features are chosen according to their influence, which means those from all the dataset with the k-highest MI values.

Following the same criteria, Figure \ref{fig:MASE_evol_sources} shows the evolution of the MASE, albeit, in this case, considering the source of the inputs (i.e., UE or BS). 
Noted that each figure is represented with a different y-axis in order to ease the analysis of the results.

\subsubsection{Latency}

Focusing on Figure \ref{fig:latency_mase}, it can be observed that all regression techniques generally achieve MASE values below 1.
Here, the LR technique, which is taken in this work as a baseline, can reach a model with a MASE of 0.76 approx. Besides, its evolution indicates that only three metrics are providing information since no prediction improvement is shown by adding more metrics. This means that only these predictors present a considerable linear relationship with this KQI.
Regarding the MI values showed in Figure \ref{fig:mutualInformation}, these metrics correspond to \textit{Resolution}, \textit{fps} and \textit{PING\_avg}.

On their behalf, KRR, KNR, SVR, RF, and ANN techniques can achieve models which fit better than LR.
These techniques manage to find a relationship with more than three features, leading to improving the accuracy of the models. 
The estimation through KNR and RF models is richer while a larger amount of features is taken, although eventually it is seen as a slight worsening. In this sense, the introduction of 12 metrics allows attaining values of MASE about 0.65 approx.
However, this behaviour is not followed by the rest of the approaches.

KRR technique takes seven features to create a model with less estimation error (i.e., MASE $\approx$ 0.66). From this number of features, the KRR models' accuracy begins to degrade.
The same applies to SVR and ANN algorithms: the best performance is obtained with eight and nine features respectively (i.e., MASE $\approx$ 0.64), becoming worst as the number of inputs is increasing.

In this case, ANN is positioned as the most suitable technique for the prediction of \textit{CGlatency}.
In addition, it can be also appreciated that their use allows obtaining good estimation with a low number of features, outperforming the rest of the techniques up to the inclusion of eight predictors.

Analogously, in Figure \ref{fig:latency_mase_sources}, the evolution of MASE considering inputs from the CG server and UE (labeled as UE) shows a similar trend to before. 
This behaviour resides on the fact that, as it can be seen in Figure \ref{fig:mutualInformation}, metrics from the CG server and UE have more MI than BS metrics. Therefore, the models with few entries match.
This is  shown in the case of the LR technique: the models show the same behaviour for the cases of LR and LR(UE). As previously seen, this algorithm is not able to find strong links beyond the metrics \textit{Resolution}, \textit{fps} and \textit{PING\_avg}.

Similarly, KRR(UE), SVR(UE) and ANN(UE) models achieve closer results in comparison with previous MASE values. Nonetheless, for this latter, no betterment is shown beyond the use of three inputs. 
In their case, KNR(UE) and RF(UE) are the ones that show more difference, albeit their top values are still similar to the previous accomplished.
Hence, ANN is again maintained as the most suitable choice when only metrics from the UE are available.
 
Now, changing the focus to the BS metrics, results show a general degradation for all the cases. 
This trend is highly appreciable for the case of LR(BS), ANN(BS), and KRR(BS), whose gaps regarding the best case go up to approximately 0.2, 0.1, and 0.05, respectively. Moreover, from these latter two, it is observed an extremely similar operation.

Conversely, others such as SVR and RF are capable of achieving similar values with both UE and BS metrics. In this sense, both approaches reach a MASE around 0.67 with six predictors.
Finally, considering the source of the features, KNR can get better estimation with BS inputs rather than with UE metrics. Besides, with all the BS features, the model performs with an error close to the ANN(UE) models, which are the best obtained with UE metrics.
In this way, when there are only metrics from the CG session and the BS, KNR models can be confirmed as the best option for estimating the \textit{CGlatency}.

\subsubsection{Freeze Percent}
Figure \ref{fig:freeze_mase} depicts the evolution of MASE for this KQI. 
First, it is remarkable that the vast majority of models for this case show much better accuracy than in the case of \textit{CGlatency}. 
Noted that it has not been possible to accomplish a model with SVR that converge with the data (i.e., MASE = 4.6). So, in order to ease the analysis of the evolution of the other metrics, SVR has not been represented either in Figure \ref{fig:freeze_mase} and \ref{fig:freeze_mase_sources}.

Thus, traditional LR technique attains MASE values around 0.47. Moreover, from their evolution it is derived that four predictors have strong relationship with the KQI. Considering the MI values represented in Figure \ref{fig:mutualInformation}, these are \textit{PING\_avg}, \textit{Resolution}, \textit{RSRP} and \textit{SINR}. 

As it occurred with \textit{CGlatency}, the use of advanced techniques such as KRR, KNR, RF, and ANN allows obtaining a considerable improvement of the prediction's accuracy. 
Among them, even though all models are considered excellent, KRR is positioned as the worst technique, obtaining a quite good MASE with values around 0.23. This value is reached when five inputs are taken, degrading the performance from this point at the same time that the number of features grows.

Considering ANN models, the best results (i.e., MASE $\approx$ 0.16) are obtained with nine inputs, getting worse when more than 11 features are introduced.
A slight improvement is achieved with the KNR technique. Besides, KNR results are accomplished with fewer predictors (i.e., $N_{feat}$ = 5) than in the case of ANN.
Nonetheless, these results are demoted when more than eight features are considered, being the prediction error even a bit superior to the obtained in ANN models.

However, the best prediction is reached with RF models. Their use guarantees the minor estimation error for each number of inputs, except for $N_{feat}$ = 3.
Additionally, there is only a subtle difference between the value reached with five metrics and the best one, which is achieved with eight features.
%Thus, for the case in which there are metrics from both the UE and BS, it can be concluded that RF is the best option.

Considering the source of the metrics, again it is appreciated that the use of data from the UE provides similar results to a mixed-use.
Like in \textit{CGlatency}, this behaviour is due to the fact that the most influential parameters for this KQI come from the CG session and UE.

Thus, Figure \ref{fig:freeze_mase_sources} shows similar results than the described before.
On the one hand, LR(UE) models are able to halve the estimation error of a naive prediction (i.e., MASE $\approx$ 0.5). Unlike before, this value is now obtained with five predictors, keeping constant beyond the use of five predictors.

On the other hand, these results are improved by the other regressions.
Thus, KRR(UE)  and ANN(UE) show the same performance as before, albeit their MASE becomes gradually demoted when more than five inputs are used.
Moreover, RF(UE) still performs with the best MASE, but in this case, the minimum is reached with only four predictors.
Here, KNR(UE) manages the same behaviour than RF(UE) for $N_{feat} \leq 4$. However, it is observed that beyond this number of features, their performance begins to become gradually worse while the amount of features grows.
Hence, both KNR(UE) and RF(UE) can be considered adequate for the prediction of \textit{FreezePercent} with only data from the UE.

When it comes to metrics from BS, LR(BS) gets values above 0.82.
Regarding the other techniques, these results are considered poor. They show values far from the best achievement (i.e., MASE $\approx$ 0.15 achieved by KNR (UE) and RF (UE)), and at the same time that closer to a naive forecast.

A similar verdict can be given for KRR(BS) and ANN(BS). Even the built models can be considered quite good in terms of MASE, the existing gap that they have with other techniques leads to their discarding.

In this scope, as seen in Figure \ref{fig:freeze_mase_sources}, RF(BS) and KNR(BS) are the unique techniques that can be highlighted as acceptable. Nonetheless, the precision provided by KNR(BS) is better than RF(BS).

Thus, it can be concluded that KNR is the most reasonable technique to use when there are metrics available from only one part of the architecture. Otherwise, the best estimation of \textit{FreezePercent} would be achieved through RF models.

\subsubsection{EFPS}

The effectiveness of each technique for the estimation of this KQI is assessed through Figure \ref{fig:efps_mase}.

Firstly, it is appreciated that this KQI has a stronger linear relationship with some inputs than for the case of \textit{CGlatency} and \textit{Freeze Percent}.
In this way, the best MASE values for the traditional LR are located around 0.32. Besides, it is perceived that the biggest improvement is obtained with two features.
Regarding the results of MI information represented in Figure \ref{fig:mutualInformation}, these are particularly the configured frame rate in the client (i.e., \textit{fps} parameter) and the average ping  between the UE and the BS (i.e., \textit{PING\_avg}). 

Following the trend previously seen for the other KQIs, a better estimation is accomplished with the advanced techniques, which are able to establish powerful relationships between the metrics.
Thus, KRR and SVR exhibit a similar evolution with less than five features, although the latter attains better values.
From this point, KRR begins to follow a negative tendency, meanwhile, SVR models still improve up to the consideration of seven predictors.

When it comes to ANN, models perform with smaller mistakes regarding the previous techniques. This algorithm takes advantage of 11 of the 13 available metrics to achieve the lowest error (i.e., MASE $\approx$ 0.1).
Similar values are also attained by RF and KNR techniques. However, they are able to get with fewer features than ANN. 
On the one hand, RF takes seven features to accomplish the best model. After that, no considerable variations appear.
On the other hand, KNR needs only four predictors to provide the best estimation. Later, some fluctuations are perceived, although eventually, MASE becomes again closer to the previous values.

Conversely, Figure \ref{fig:efps_mase_sources} represents the evolution of the MASE for the \textit{EFPS} models.
Here, it is appraised analogous values as described above. In this way, LR(UE), KRR(UE) and SVR(UE) keep the same trend as before. 
The same occurs for RF(UE) models, which maintain their behaviour even if only UE metrics are considered.
Lastly, for the situation of ANN(UE), and KNR(UE), it is appreciated a gradual demotion of the MASE when more than five inputs are taken. Nonetheless, the best values in these situations match with the previously obtained.

Focusing on the use of metrics from the BS, it is perceived as a degradation for the LR(BS) and KRR(BS). 
The models are not able to perform with a MASE below 0.40 and 0.23 respectively, which supposes a difference of around 0.1 with the best results accomplished by these techniques. 
Beyond them, SVR(BS), ANN(BS), and RF(BS) also show degradation of their MASE regarding the use of UE metrics. Nonetheless, these differences are below 0.5 in terms of MASE. 
Finally, KNR(BS) follows the same course as the other KQIs.
Once all the BS metrics are used, the model that is based in KNR is able to minimise the prediction error, being their values similar to those obtained with UE metrics.

Hence, it is determined that KNR is the best technique for the forecasting of \textit{EFPS} since it performs the best prediction in any of the three cases analysed. Besides, it is also remarkable that the vast majority of considered models attain a substantially good estimation for this KQI, regardless of the sources of the predictors.

\iffalse
\begin{figure}[h]
\centering
\includegraphics[width=0.9 \columnwidth]{images/cg_latency50Percent_MASE.pdf}
\caption{MASE evolution for Cloud Gaming }%\vspace{-1\baselineskip}
\label{fig:MASE_latency}
\vspace*{-\baselineskip}
\end{figure} 

\begin{figure}[h]
\centering
\includegraphics[width=0.9 \columnwidth]{images/cg_latency50Percent_MASE_sources.pdf}
\caption{MASE evolution for Cloud Gaming with different sources}%\vspace{-1\baselineskip}
\label{fig:MASE_latency_sources}
\vspace*{-\baselineskip}
\end{figure}

\subsubsection{Freeze Percent}

\begin{figure}[h]
\centering
\includegraphics[width=0.9 \columnwidth]{images/freezePercent_MASE.pdf}
\caption{MASE evolution for Freeze Percent}%\vspace{-1\baselineskip}
\label{fig:MASE_freeze}
\vspace*{-\baselineskip}
\end{figure} 

\begin{figure}[h]
\centering
\includegraphics[width=0.9 \columnwidth]{images/freezePercent_MASE_sources.pdf}
\caption{MASE evolution for Freeze Percent with different sources}%\vspace{-1\baselineskip}
\label{fig:MASE_freeze_sources}
\vspace*{-\baselineskip}
\end{figure} 

\subsubsection{EFPS}

\begin{figure}[h]
\centering
\includegraphics[width=0.9 \columnwidth]{images/efps_MASE.pdf}
\caption{MASE evolution for efps}%\vspace{-1\baselineskip}
\label{fig:MASE_efps}
\vspace*{-\baselineskip}
\end{figure} 

\begin{figure}[h]
\centering
\includegraphics[width=0.9 \columnwidth]{images/efps_MASE_sources.pdf}
\caption{MASE evolution for efps with different sources}%\vspace{-1\baselineskip}
\label{fig:MASE_efps_sources}
\vspace*{-\baselineskip}
\end{figure} 

\fi

\subsection{Prediction time}
In addition to the estimation error, models are also evaluated through their prediction time, which means the time that each model takes to estimate values. This assessment has been carried out in an Intel Xeon Silver with 12 CPU cores, a 2.2 GHz clock frequency, and 128 GB of RAM.

Table \ref{tab:executionTimes} summarises the mean execution time (in milliseconds) of 1000 predictions of each KQI by the use of different regression techniques.
For simplicity, it has been considered models that take all the features from the dataset (13 features), as well as the ones that attain the best prediction in terms of MASE. 

Thus, as expected, models based on LR technique are the fastest thanks to its simplicity, requiring less than 1ms in any case.

When it comes to more advanced techniques, these values become much higher. KNR needs around 17ms to get an estimation for both \textit{FreezePercent} and \textit{EFPS}, being double for the case of \textit{CGlatency}. 

These values soar for the case of SVR, which takes around 47 ms and 31 ms to predict \textit{CGlatency} and \textit{EFPS} respectively. Note that SVR-based models were not able to be trained for \textit{FreezePercent}. 
Similarly, prediction times of KRR are around 53 ms regardless of the KQI to estimate.
On their behalf, ANN models spend up to 4 times as much time providing an estimation. This huge increment resides on the high complexity of these models.

Finally, among these advanced techniques, RF takes the smaller time for latency forecasting (i.e., 11.69 ms). However, it shows similar prediction times to KNR for the case of \textit{FreezePercent} and \textit{EFPS}

Hence, it can be concluded that models based on RF and KNR is less computationally demanding, albeit the existing differences with the other techniques can be despised since predictors rarely have refresh periods below 1 second. 

\begin{table}[]
\centering
\caption{Model's execution times (ms)}
\label{tab:executionTimes}
\begin{tabular}{@{}cccc@{}}
\toprule
\textbf{Technique}    & \textbf{CGLatency} & \textbf{Freeze Percent} & \textbf{EFPS} \\ \midrule
\textbf{LR}  & 0.079              & 0.086                   & 0.079         \\ \midrule
\textbf{KNR} & 30.89              & 16.84                   & 17.03         \\ \midrule
\textbf{SVR} & 46.94              & -                       & 31.37         \\ \midrule
\textbf{KRR} & 52.96              & 53.52                   & 52.98         \\ \midrule
\textbf{RF}  & 11.69              & 16.07                   & 20.59         \\ \midrule
\textbf{ANN} & 208.23             & 234.09                  & 165.6         \\ \bottomrule
\end{tabular}

\end{table}

\section{Conclusions}
\label{sec:Conclusions}

The Cloud Gaming (CG) paradigm is becoming one of the most promising alternatives to traditional gaming. 
Thanks to its architecture, users can play realistic games at any time, at any place, and regardless of the computational capabilities of their devices. 
In exchange, this paradigm puts more pressure on the network, triggering that the service's QoE highly depends on the user's connectivity. 

In this context, the present work has proposed a methodology for evaluating the CG QoE through the obtaining of its KQIs: \textit{CGlatency}, \textit{FreezePercent}, and \textit{EFPS}.
Besides, a comprehensive study of several Machine Learning techniques has been carried out in order to estimate such KQIs from session configuration, UE, and network metrics.
Here, traditional linear regression has been compared with four of the most extended ML techniques.

The obtained results prove the capability to predict the proposed KQIs by using readily available metrics from  the configuration of the CG session, the UE, and the BS. These relationships, which are presented in a stronger way for the case of \textit{FreezePercent} and \textit{EFPS}, allow performing satisfactory predictions in any case (i.e., MASE $< 1$).

Thus, classical LR models attain satisfactory results for \textit{FreezePercent} and \textit{EFPS} (i.e., MASE = 0.47 and 0.32 respectively). For the case of \textit{CGlatency}, the prediction is less accurate, albeit still acceptable (i.e., MASE = 0.76).
%Nevertheless, these results undergo a significant improvement when more advanced techniques are applied. Thanks to the search for non-linear relationships, the use of RF, KNR and, ANN models provides high accuracy prediction of these KQIs.
Nevertheless, these results undergo a significant improvement when more advanced techniques are applied. Thanks to the search for non-linear relationships, \textit{CGlatency} can be estimated through ANN models with a MASE of 0.63 approx.
Along the same lines, the use of RF and KNR models drives the prediction accuracy of \textit{Freeze Percent} and \textit{EFPS} to MASE values of 0.12 and 0.08 approximately.

These values are similar when only UE metrics are introduced to the models. 
However, when it comes to the use of only BS metrics, it is seen a general demotion of the models' quality, with the exception of KNR. By using all the available BS metrics of the dataset, this approach obtains estimations close to the best achieved in the other cases.

Focusing on the accuracy and execution time compromise, RF has been positioned as one of the most interesting techniques to use. In addition to the small prediction error obtained, it only needs 20~ms to estimate any KQI.
On its behalf, KNR has similar behaviour when \textit{FreezePercent} and \textit{EFPS} are estimated. Nonetheless, its performance versatility regarding the source of the metrics triggers to highly consider its use for estimating the KQI of CG services.

Future works will focus on the introduction of this approach in the Open RAN paradigm: the RAN Intelligent Controller (\textit{RIC}) is positioned as the main element for the managing of the cellular network's RAN.  Thus, the development of applications for the Non-Real-Time and Near Real-Time RIC (defined as rApps and xApps respectively) will ease the implementation of the presented approach, as well as support the provision of CG services in the next-generation networks.

\iffalse
\appendices

Appendixes, if needed, appear before the acknowledgment.

\fi

%This work has been partially funded by the ``Ministerio de Asuntos Econ\'omico y transformaci\'on digital'' and ERDF - European Regional Development Fund (red.es, ``Piloto 5G Andaluc\'ia, Caso 31 Estudio OpenRAN''), Junta de Andaluc\'ia and ERDF (Programa Operativo FEDER Andaluc\'ia 2014-2020) project IDADE-5G (UMA18-FEDERJA-201) and by "Ministerio de ciencia, innovaci\'on y universidades" under grant agreement FPU19/04468.

\bibliographystyle{IEEEtran}

\bibliography{Bibliography} 

\begin{thebibliography}{10}
\providecommand{\url}[1]{#1}
\csname url@samestyle\endcsname
\providecommand{\newblock}{\relax}
\providecommand{\bibinfo}[2]{#2}
\providecommand{\BIBentrySTDinterwordspacing}{\spaceskip=0pt\relax}
\providecommand{\BIBentryALTinterwordstretchfactor}{4}
\providecommand{\BIBentryALTinterwordspacing}{\spaceskip=\fontdimen2\font plus
\BIBentryALTinterwordstretchfactor\fontdimen3\font minus
  \fontdimen4\font\relax}
\providecommand{\BIBforeignlanguage}[2]{{%
\expandafter\ifx\csname l@#1\endcsname\relax
\typeout{** WARNING: IEEEtran.bst: No hyphenation pattern has been}%
\typeout{** loaded for the language `#1'. Using the pattern for}%
\typeout{** the default language instead.}%
\else
\language=\csname l@#1\endcsname
\fi
#2}}
\providecommand{\BIBdecl}{\relax}
\BIBdecl

\bibitem{MARCHAND2013141}
\BIBentryALTinterwordspacing
A.~Marchand and T.~Hennig-Thurau, ``Value creation in the video game industry:
  Industry economics, consumer benefits, and research opportunities,''
  \emph{Journal of Interactive Marketing}, vol.~27, no.~3, pp. 141--157, 2013.
  [Online]. Available:
  \url{https://www.sciencedirect.com/science/article/pii/S1094996813000170}
\BIBentrySTDinterwordspacing

\bibitem{Ducheneaut}
\BIBentryALTinterwordspacing
N.~Ducheneaut and R.~J. Moore, ``The social side of gaming: A study of
  interaction patterns in a massively multiplayer online game,'' in
  \emph{Proceedings of the 2004 ACM Conference on Computer Supported
  Cooperative Work}, ser. CSCW '04.\hskip 1em plus 0.5em minus 0.4em\relax New
  York, NY, USA: Association for Computing Machinery, 2004, p. 360–369.
  [Online]. Available: \url{https://doi.org/10.1145/1031607.1031667}
\BIBentrySTDinterwordspacing

\bibitem{SZELL2010313}
\BIBentryALTinterwordspacing
M.~Szell and S.~Thurner, ``Measuring social dynamics in a massive multiplayer
  online game,'' \emph{Social Networks}, vol.~32, no.~4, pp. 313--329, 2010.
  [Online]. Available:
  \url{https://www.sciencedirect.com/science/article/pii/S0378873310000316}
\BIBentrySTDinterwordspacing

\bibitem{cai2016survey}
W.~Cai, R.~Shea, C.-Y. Huang, K.-T. Chen, J.~Liu, V.~C. Leung, and C.-H. Hsu,
  ``A survey on cloud gaming: Future of computer games,'' \emph{IEEE Access},
  vol.~4, pp. 7605--7620, 2016.

\bibitem{Abdallah2018}
\BIBentryALTinterwordspacing
M.~Abdallah, C.~Griwodz, K.-T. Chen, G.~Simon, P.-C. Wang, and C.-H. Hsu,
  ``Delay-sensitive video computing in the cloud: A survey,'' \emph{ACM Trans.
  Multimedia Comput. Commun. Appl.}, vol.~14, no.~3s, Jun. 2018. [Online].
  Available: \url{https://doi.org/10.1145/3212804}
\BIBentrySTDinterwordspacing

\bibitem{BaenaCG2022}
\BIBentryALTinterwordspacing
C.~Baena, ``{{Gaming in the Cloud: 5G as the pillar for future gaming
  approaches}},'' 12 2022. [Online]. Available:
  \url{https://www.techrxiv.org/articles/preprint/Gaming_in_the_Cloud_5G_as_the_pillar_for_future_gaming_approaches/21665645}
\BIBentrySTDinterwordspacing

\bibitem{playstation}
\BIBentryALTinterwordspacing
``{PS Now: On-demand PlayStation games on PS5, PS4 or PC}.'' [Online].
  Available: \url{https://www.playstation.com/en-us/ps-now/}
\BIBentrySTDinterwordspacing

\bibitem{geforcenow}
\BIBentryALTinterwordspacing
``{GeForce now Cloud Gaming}.'' [Online]. Available:
  \url{https://www.nvidia.com/en-us/geforce-now/}
\BIBentrySTDinterwordspacing

\bibitem{amazonLuna}
\BIBentryALTinterwordspacing
``{Play your favorite games straight from the cloud with Amazon Luna},'' 1970.
  [Online]. Available: \url{https://www.amazon.com/luna/landing-page}
\BIBentrySTDinterwordspacing

\bibitem{xbox}
\BIBentryALTinterwordspacing
``Xbox cloud gaming with xbox game pass.'' [Online]. Available:
  \url{https://www.xbox.com/en-GB/xbox-game-pass/cloud-gaming}
\BIBentrySTDinterwordspacing

\bibitem{MLSON}
P.~V. {Klaine}, M.~A. {Imran}, O.~{Onireti}, and R.~D. {Souza}, ``{A Survey of
  Machine Learning Techniques Applied to Self-Organizing Cellular Networks},''
  \emph{IEEE Communications Surveys Tutorials}, vol.~19, no.~4, pp. 2392--2431,
  2017.

\bibitem{UMLNetworking}
M.~Usama, J.~Qadir, A.~Raza, H.~Arif, K.-L. Yau, Y.~Elkhatib, A.~Hussain, and
  A.~Al-Fuqaha, ``{Unsupervised Machine Learning for Networking: Techniques,
  Applications and Research Challenges},'' \emph{IEEE Access}, vol.~PP, 09
  2017.

\bibitem{Palacios2018}
D.~Palacios, S.~Fortes, I.~de-la Bandera, and R.~Barco, ``{Self-Healing
  Framework for Next-Generation Networks through Dimensionality Reduction},''
  \emph{IEEE Communications Magazine}, vol.~56, no.~7, pp. 170--176, 2018.

\bibitem{LocationAwareness}
\BIBentryALTinterwordspacing
S.~Fortes, C.~Baena, J.~Villegas, E.~Baena, M.~Z. Asghar, and R.~Barco,
  ``{Location-Awareness for Failure Management in Cellular Networks: An
  Integrated Approach},'' \emph{Sensors}, vol.~21, no.~4, 2021. [Online].
  Available: \url{https://www.mdpi.com/1424-8220/21/4/1501}
\BIBentrySTDinterwordspacing

\bibitem{herrera2019modeling}
A.~Herrera-Garcia, S.~Fortes, E.~Baena, J.~Mendoza, C.~Baena, and R.~Barco,
  ``{Modeling of key quality indicators for end-to-end network management:
  Preparing for 5G},'' \emph{IEEE Vehicular Technology Magazine}, vol.~14,
  no.~4, pp. 76--84, 2019.

\bibitem{baena2020estimation}
C.~Baena, S.~Fortes, E.~Baena, and R.~Barco, ``{Estimation of Video Streaming
  KQIs for Radio Access Negotiation in Network Slicing Scenarios},'' \emph{IEEE
  Communications Letters}, vol.~24, no.~6, pp. 1304--1307, 2020.

\bibitem{jarschel2013gaming}
M.~Jarschel, D.~Schlosser, S.~Scheuring, and T.~Ho{\ss}feld, ``Gaming in the
  clouds: Qoe and the users’ perspective,'' \emph{Mathematical and Computer
  Modelling}, vol.~57, no. 11-12, pp. 2883--2894, 2013.

\bibitem{claypool2014effects}
M.~Claypool and D.~Finkel, ``The effects of latency on player performance in
  cloud-based games,'' in \emph{2014 13th Annual Workshop on Network and
  Systems Support for Games}.\hskip 1em plus 0.5em minus 0.4em\relax IEEE,
  2014, pp. 1--6.

\bibitem{Raaen2014}
K.~{Raaen}, R.~{Eg}, and C.~{Griwodz}, ``Can gamers detect cloud delay?'' in
  \emph{2014 13th Annual Workshop on Network and Systems Support for Games},
  2014, pp. 1--3.

\bibitem{Sabet2019DelayVariation}
S.~S. {Sabet}, S.~{Schmidt}, C.~{Griwodz}, and S.~{Möller}, ``Towards the
  impact of gamers' adaptation to delay variation on gaming quality of
  experience,'' in \emph{2019 Eleventh International Conference on Quality of
  Multimedia Experience (QoMEX)}, 2019, pp. 1--6.

\bibitem{Sabet2020}
Sabet and other, ``{Towards the Impact of Gamers Strategy and User Inputs on
  the Delay Sensitivity of Cloud Games},'' in \emph{International Conference on
  Quality of Multimedia Experience}.\hskip 1em plus 0.5em minus 0.4em\relax IEE
  Inc., may 2020.

\bibitem{claypool2006latency}
M.~Claypool and K.~Claypool, ``Latency and player actions in online games,''
  \emph{Communications of the ACM}, vol.~49, no.~11, pp. 40--45, 2006.

\bibitem{quax2013evaluation}
P.~Quax, A.~Beznosyk, W.~Vanmontfort, R.~Marx, and W.~Lamotte, ``An evaluation
  of the impact of game genre on user experience in cloud gaming,'' in
  \emph{2013 IEEE International Games Innovation Conference (IGIC)}.\hskip 1em
  plus 0.5em minus 0.4em\relax IEEE, 2013, pp. 216--221.

\bibitem{claypool2010lat}
M.~Claypool and K.~Claypool, ``{Latency can kill: precision and deadline in
  online games},'' \emph{Proceedings of the first annual ACM SIGMM conference
  on Multimedia systems}, pp. 215--222, 2010.

\bibitem{lee2012all}
Y.-T. Lee, K.-T. Chen, H.-I. Su, and C.-L. Lei, ``Are all games equally
  cloud-gaming-friendly? an electromyographic approach,'' in \emph{2012 11th
  Annual Workshop on Network and Systems Support for Games (NetGames)}.\hskip
  1em plus 0.5em minus 0.4em\relax IEEE, 2012, pp. 1--6.

\bibitem{Sabet2020DelaySensityClassification}
\BIBentryALTinterwordspacing
S.~S. Sabet, S.~Schmidt, S.~Zadtootaghaj, C.~Griwodz, and S.~M\"{o}ller,
  ``{Delay Sensitivity Classification of Cloud Gaming Content},'' in
  \emph{Proceedings of the 12th ACM International Workshop on Immersive Mixed
  and Virtual Environment Systems}, ser. MMVE '20.\hskip 1em plus 0.5em minus
  0.4em\relax New York, NY, USA: Association for Computing Machinery, 2020, p.
  25–30. [Online]. Available: \url{https://doi.org/10.1145/3386293.3397116}
\BIBentrySTDinterwordspacing

\bibitem{slivar2015}
I.~Slivar, M.~Suznjevic, L.~Skorin-Kapov, and M.~Matijasevic, ``Empirical qoe
  study of in-home streaming of online games,'' \emph{Annual Workshop on
  Network and Systems Support for Games}, vol. 2015, 01 2015.

\bibitem{moller2013factors}
S.~M{\"o}ller, D.~Pommer, J.~Beyer, and J.~Rake-Revelant, ``Factors influencing
  gaming qoe: Lessons learned from the evaluation of cloud gaming services,''
  in \emph{Proceedings of the 4th International Workshop on Perceptual Quality
  of Systems (PQS 2013)}, 2013, pp. 1--5.

\bibitem{chang2011understanding}
Y.-C. Chang, P.-H. Tseng, K.-T. Chen, and C.-L. Lei, ``Understanding the
  performance of thin-client gaming,'' in \emph{2011 IEEE International
  Workshop Technical Committee on Communications Quality and Reliability
  (CQR)}.\hskip 1em plus 0.5em minus 0.4em\relax IEEE, 2011, pp. 1--6.

\bibitem{Claypool}
M.~{Claypool}, D.~{Finkel}, A.~{Grant}, and M.~{Solano}, ``Thin to win? network
  performance analysis of the onlive thin client game system,'' in \emph{2012
  11th Annual Workshop on Network and Systems Support for Games (NetGames)},
  2012, pp. 1--6.

\bibitem{penaherrera2021measuring}
O.~S. Pe{\~n}aherrera-Pulla, C.~Baena, S.~Fortes, E.~Baena, and R.~Barco,
  ``{Measuring Key Quality Indicators in Cloud Gaming: Framework and Assessment
  Over Wireless Networks},'' \emph{Sensors}, vol.~21, no.~4, p. 1387, 2021.

\bibitem{moonlight}
\BIBentryALTinterwordspacing
``{Moonlight. An open source Nvidia gamestream client}.''\hskip 1em plus 0.5em
  minus 0.4em\relax Nvidia, Accessed July 2021. [Online]. Available:
  \url{https://moonlight-stream.org/}
\BIBentrySTDinterwordspacing

\bibitem{BaenaDataset2022}
\BIBentryALTinterwordspacing
C.~Baena, O.~Peñaherrera, L.~Camacho, R.~Barco, and S.~Fortes, ``{{Video
  Streaming and Cloud Gaming services over 4G and 5G: a complete network and
  service metrics dataset}},'' 11 2022. [Online]. Available:
  \url{https://www.techrxiv.org/articles/preprint/Video_Streaming_and_Cloud_Gaming_services_over_4G_and_5G_a_complete_network_and_service_metrics_dataset/21456267}
\BIBentrySTDinterwordspacing

\bibitem{frameworkCrowd}
C.~Baena, S.~Fortes, O.~Peñaherrera, and R.~Barco, ``{A Framework to boost the
  potential of network-in-a-box solutions},'' in \emph{2021 12th International
  Conference on Network of the Future (NoF)}, 2021, pp. 1--3.

\bibitem{MASEref}
R.~Hyndman and A.~Koehler, ``Another look at measures of forecast accuracy,''
  \emph{International Journal of Forecasting}, vol.~22, pp. 679--688, 02 2006.

\end{thebibliography}
% This section is the exampled bibliography 

\end{document}